\documentclass[lettersize,journal]{IEEEtran}
\usepackage{amsmath,amsfonts}
\usepackage{array}
\usepackage[caption=false,font=normalsize,labelfont=sf,textfont=sf]{subfig}
\usepackage{textcomp}
\usepackage{stfloats}
\usepackage{url}
\usepackage{verbatim}
\usepackage{graphicx}
\usepackage{hyperref}
\usepackage{enumitem}
\usepackage{multirow}
\usepackage{placeins}
\usepackage{booktabs}
\usepackage{algorithm}
\usepackage[noend]{algpseudocode}
\usepackage{prettyref}
\DeclareMathOperator*{\argmin}{arg\,min}
\algnewcommand{\LineComment}[1]{\State \(\triangleright\) #1}
\usepackage{mathtools}
\usepackage{bm}

\hypersetup{
    colorlinks=true, 
    linkcolor=blue,  
    citecolor=blue,  
    urlcolor=blue    
}

\usepackage{orcidlink}

\newcommand{\timeSet}{\mathcal{T}}

\def\BibTeX{{\rm B\kern-.05em{\sc i\kern-.025em b}\kern-.08em
    T\kern-.1667em\lower.7ex\hbox{E}\kern-.125emX}}
\usepackage{balance}
\begin{document}
\title{Simultaneous Trajectory Optimization and Contact Selection for Contact-rich Manipulation with High-Fidelity Geometry}
\author{Mengchao Zhang, Devesh K. Jha, Arvind U. Raghunathan, Kris Hauser

\thanks{Mengchao Zhang is with the Department of Mechanical Science and Engineering, University of Illinois at Urbana-Champaign, Urbana, IL, 61801, USA. (e-mail: mz17@illinois.edu)}
\thanks{Devesh K. Jha and Arvind U. Raghunathan are with Mitsubishi Electric Research Laboratories (MERL), Cambridge, MA, 02139, USA. (e-mail:\{jha,raghunathan\}@merl.com)}
\thanks{Kris Hauser is with the Department of Computer Science, University of Illinois at Urbana-Champaign, Urbana, IL, 61801, USA. (e-mail: kkhauser@illinois.edu)}
\thanks{Code is available at: \url{https://github.com/zmccmzty/STOCS-3D}.}}



\maketitle

\begin{abstract}
Contact-implicit trajectory optimization (CITO) is an effective method to plan complex trajectories for various contact-rich systems including manipulation and locomotion. CITO formulates a mathematical program with complementarity constraints (MPCC) that enforces that contact forces must be zero when points are not in contact. However, MPCC solve times increase steeply with the number of allowable points of contact, which limits CITO's applicability to problems in which only a few, simple geometries are allowed to make contact. This paper introduces simultaneous trajectory optimization and contact selection (STOCS), as an extension of CITO that overcomes this limitation. The innovation of STOCS is to identify salient contact points and times inside the iterative trajectory optimization process. This effectively reduces the number of variables and constraints in each MPCC invocation. The STOCS framework, instantiated with key contact identification subroutines, renders the optimization of manipulation trajectories  computationally tractable even for high-fidelity geometries consisting of tens of thousands of vertices. 


\end{abstract}


\section{Introduction}\label{sec:introduction}

\IEEEPARstart{H}{umans} and other organisms treat contact as a fact of life and utilize contact to perform dexterous manipulation of objects and agile locomotion. In contrast, the majority of current robots avoid making contact with objects as much as possible, and tend to avoid contact-rich manipulations like pushing, sliding, and rolling \cite{dogar2012physics,eppner2015planning}. 
Trajectory optimization \cite{kelly2017introduction} has been investigated as a tool for generating high quality manipulations, but choosing an effective mathematical representation of making and breaking contact remains a major research challenge. Two general classes of methods are available: hybrid trajectory optimization and contact-implicit trajectory optimization (CITO). Hybrid trajectory optimization divides a trajectory into segments in which the set of contacts remains constant, but it requires the contact mode sequence to be known in advance~\cite{schultz2009modeling} or explored by an auxiliary discrete search. CITO \cite{posa2014direct,mordatch2012contact,mordatch2012discovery,patel2019contact, onol2020tuning} allows the optimizer to choose the sequence of contact within the optimization loop. CITO formulates contact as a complementarity constraint to ensure that the contact forces can be non-zero if and only if a point is in contact~\cite{posa2014direct}. Although the resulting mathematical programming with complementary constraint (MPCC) \cite{luo1996mathematical} formulation is less restrictive than hybrid trajectory optimization, it still requires a set of predefined allowable contact points on the object. Moreover, MPCC rapidly becomes more challenging to solve as the number of complementarity constraints increases, so past CITO applications were limited to a small handful of potential contact points.

\begin{figure*}
    \centering
    \includegraphics[width=1\linewidth]{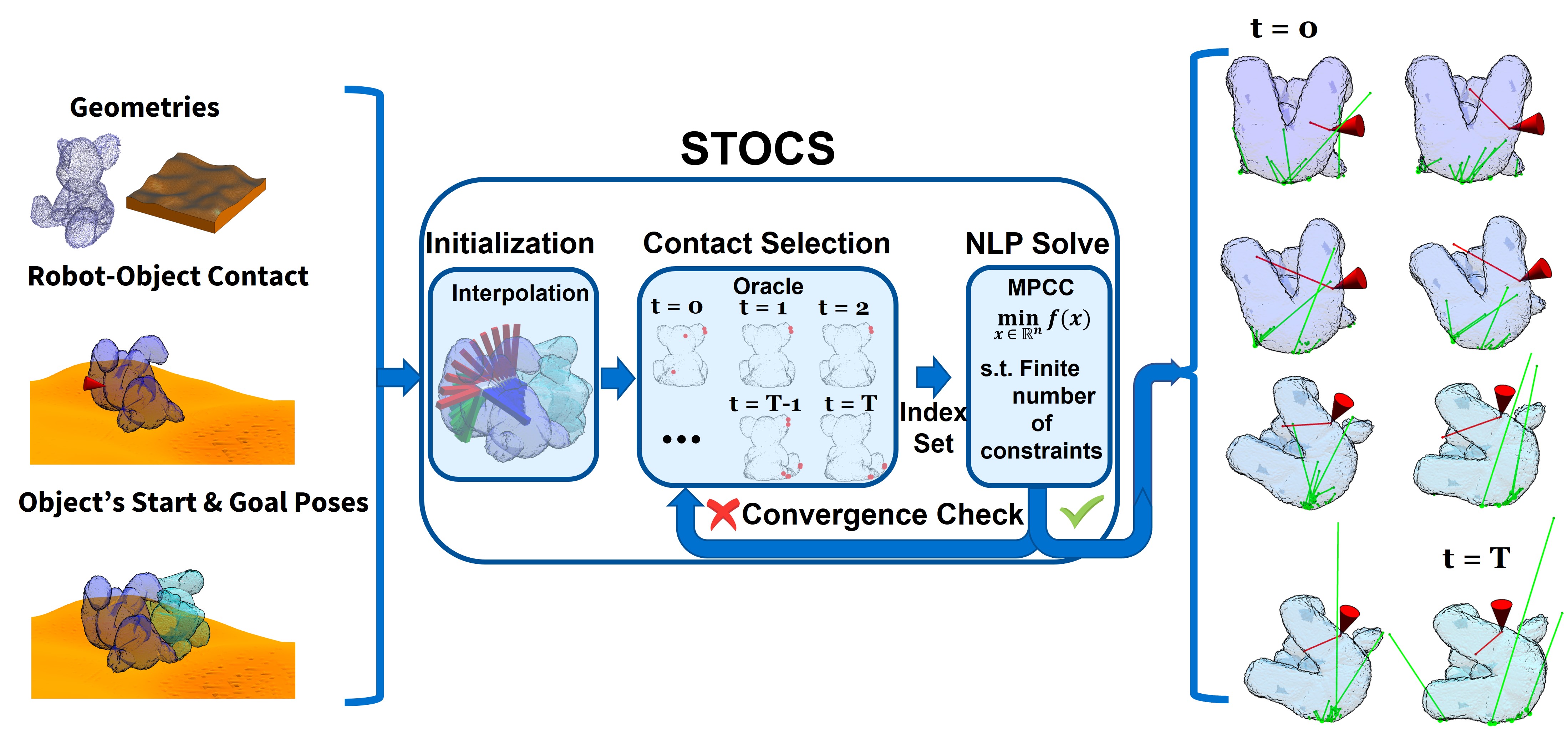}\\
    \caption{STOCS accepts as input the high-fidelity geometry of the object (represented by a dense point cloud) and the environment (represented by a signed distance field), the robot's contact point, and start and goal poses of the object (left). The STOCS algorithm first generates an initial trajectory by linearly interpolating between the start and goal poses, and then it iterates between selecting contact points and solving a finite-dimensional MPCC to decide a step direction until the convergence criteria are met (center). As output (right), STOCS produces the pose of the object, active object-environment contact points (green dots) and forces (green lines), and manipulation force (red lines).  Nonpenetration, Coulomb friction, complementarity, and quasi-dynamic stability are enforced throughout the trajectory. [Best viewed in color.]}
    \label{fig:exp_setup}
\end{figure*}

This paper introduces the simultaneous trajectory optimization and contact selection (STOCS) algorithm to address the scaling problem in contact-implicit trajectory optimization. To the best of our knowledge, this is the first method capable of optimizing contact-rich manipulation trajectories with high-fidelity geometric representations in 3D. This method paves the way for manipulation planning with raw sensor input, such as point clouds derived from RGBD images, and eliminates the need for geometry simplification.

STOCS applies an infinite programming (IP) approach to dynamically instantiate possible contact points and contact times between the object and environment inside the optimization loop, and hence the resulting MPCCs become far more tractable to solve. The primary contribution of this work is to extend prior work on IP for robot pose optimization~\cite{zhang2021semi} to the trajectory optimization setting. This paper describes methods for encoding quasi-dynamic constraints in trajectory optimization, and also presents a novel method for selecting salient contact points and contact times.  This method, Time-active Maximum Violation Oracle (TAMVO) with spatial disturbance and temporal smoothing, encourages the IP framework to converge quickly toward a feasible solution.  we demonstrate the effectiveness and efficiency of STOCS in solving 2D and 3D sliding, pivoting, and peg in hole tasks with irregular objects and environments, whose models can include up to tens of thousands of vertices. Without STOCS, CITO methods take hundreds to thousands of times longer, even for problems of moderate size (a few hundred vertices).

\section{Related Work}\label{sec:related_work}

Model-based trajectory planning methods have been extensively studied for contact-rich manipulation problems. The presence of contact presents significant challenges for optimization, primarily due to the stiff and non-smooth nature of contact dynamics~\cite{clarke1990optimization}. Various approaches have been proposed to address this challenge. 

{\em Hybrid trajectory optimization} approaches divide a trajectory into segments in which the contact mode (set of active contacts) remains constant \cite{harada2006natural}. This segmentation allows trajectory optimization to be cast as a large non-linear program that can be solved to optimize the timings and variables associated with each of the individual modes~\cite{chavan2020planar, 9811812}. However, manually determining an appropriate sequence of contact modes is intractable in all but the simplest problems.

Rather than predefining contact mode sequences, {\em mode search} methods~\cite{cheng2021contact,cheng2022contact} use automatically enumerate contact modes \cite{huang2021efficient} to guide the tree expansion within a sampling-based motion planning algorithm. Although this overcomes the limitations of mode sequencing, the complexity of enumerating all 3D contact modes for one
object is $\mathcal{O}(N^d)$, where $N$ represents the number of contacts and $d$ denotes the object's effective degrees of freedom, which is 3 in 2D and 6 in 3D. This superlinear growth in $N$ makes these methods impractical for high-fidelity geometric representations.

Contact-implicit trajectory optimization (CITO) offers an alternative to predefining or searching over the contact mode sequence~\cite{posa2014direct,mordatch2012contact,mordatch2012discovery}. CITO models all possible contacts with complementarity constraints and casts trajectory optimization as a mathematical program with complementarity constraints
(MPCC). These constraints allow forces to be nonzero if and only if the distance between a contact and the opposing object is zero. Given a set of potential allowable contacts, the optimizer loosens the complementarity constraints to let forces be applied at a distance, and then progressively tightens them. Through smoothing the contact dynamics using the above strategy, CITO can then simultaneously optimize the mode sequence as well as the contact forces. However, CITO becomes extremely challenging to solve when there are a large number of possible contacts, leading to large numbers of complementarity constraints. Consequently, the main limitation of this approach is that the set of allowable contacts must be predefined and needs to be relatively small.

The STOCS method introduced in this paper addresses the scaling problem of CITO by incorporating the identification of salient contact points and contact times inside trajectory optimization. This approach effectively reduces the number of variables and constraints in the resulting MPCC, rendering the computation of contact-rich manipulation trajectories for objects with complex, non-convex geometries computationally tractable.  In prior work, we introduces the infinite programming framework used here, and applied it to optimizing stable grasping poses~\cite{Zhang2021}.  In contrast, this paper extends the framework to optimize entire trajectories for contact-rich manipulations.

This article is an expanded version of a conference paper \cite{zhang2023simultaneous}. Our extensions broaden the applicability of STOCS and enables its use with more complex objects and environments. Compared to the conference version, which only worked in planar problems, the current implementation is far more efficient and we demonstrate how it can be applied to 3D problems. Moreover, we introduce a novel method for selecting contact points and contact times, called the Time-active Maximum Violation Oracle (TAMVO), that also greatly improves computational performance. This version also adds additional details, references, and experiments.
\section{Approach}\label{sec:approach}

Given a start pose and a goal pose, a trajectory that includes the object's motion and the control inputs of the manipulator needs to be planned. In this section, we describe the inputs and outputs of our method in detail, and explain how we use STOCS to solve contact-rich manipulation trajectory optimization problems. 


\subsection{Problem Description}\label{subsec:problem_description}

Our method requires the following information as inputs:
\begin{enumerate}
    \item Object initial pose region: $Q_{init} \subset SE(2) \; \text{or} \; SE(3)$.
    \item Object goal pose region: $Q_{goal} \subset SE(2) \; \text{or} \; SE(3)$.
    \item Object properties: a rigid body $\mathcal{O}$ whose geometry, mass distribution, and friction coefficients with both the environment $\mu_{env}$ and the manipulator $\mu_{mnp}$ are known.
    \item Environment properties: rigid environment $\mathcal{E}$ whose geometry is known.
    \item Robot's contact point(s) with the object: $c^{mnp}$.
    \item A time step $\Delta t$ and number of time steps $T$ in the trajectory.  
\end{enumerate}

Our method will output a trajectory $\tau$ that includes the following information at time $t$:
\begin{enumerate}
    \item Object's configuration: $q_t$.
    \item Object's velocity (angular and translational): $v_t$.
    \item Robot's contact point(s): $c^{mnp}_t$.
    \item Manipulation force: $u_t$.
    \item Object's contact points with the environment: $\tilde{Y}_t$.
    \item Contact force at each object-environment contact point: $z(y_t) \; \forall y_t \in \tilde{Y}_t$.
\end{enumerate}

In this paper, we treat all objects and environments as rigid bodies and we assume the contact between the manipulator and the object is sticking contact. Furthermore, users are provided with the flexibility to opt for either quasistatic or quasidynamic assumptions based on their specific requirements. Under the quasistatic paradigm, inertial forces are considered negligible, necessitating that the object remains in a state of force and torque equilibrium at all times. Quasidynamic manipulation accounts for scenarios where tasks may involve occasional dynamic periods, during which accelerations do not integrate into substantial velocities, and both momentum and the effects of impact restitution are negligible \cite{mason2001mechanics}.

\subsection{STOCS Trajectory Optimizer}

Overall, STOCS formulates contact-rich trajectory optimization as an {\em infinite program} (IP), which is a constrained optimization with a potentially infinite set of variables and constraints. It uses an {\em exchange method} to solve the IP, by wrapping  a contact selection outer loop around a finite optimization problem.  The inner problem formulates CITO with complementarity constraints and solves an MPCC.  In summary, the algorithm operates as follows:
\begin{enumerate}
\item Initialize the initial guess object trajectory, e.g., a linear interpolation.
\item Initialize an empty {\em candidate set} of contact points $\tilde{Y}_1,\ldots,\tilde{Y}_T$ for each time step and an empty set of contact forces.
\item Use an {\em Oracle} to identify new / removed contact points, and update $\tilde{Y}_1,\ldots,\tilde{Y}_T$.
\item Solve an MPCC for the candidate set of contact points using a small number of iterations.
\item Check for convergence.  If not converged, repeat from step~3.
\end{enumerate}
Contact forces for existing contact points are maintained from iteration to iteration to warm-start the next MPCC.  Details about the algorithm (Alg.~\ref{alg:STOCS}) and oracle design are described below.

\begin{algorithm}[tbp]
\caption{STOCS}
\label{alg:STOCS}
\begin{algorithmic}[1]
\Require $q_{start}$, $q_{goal}$, $c^{mnp}$
\State $\tilde{Y}^{0}=[\,]$ \Comment{Initialize empty constraint set}
\State $z_0 \gets \emptyset$ \Comment{Initialize empty force vector}
\State $x_0 \gets $ initialize trajectory($q_{start}$, $q_{goal}$, $c^{mnp}$)
\For{$k=1,\ldots,N^{max}$}

\LineComment{Update constraint set and guessed forces $z_{k}$}
\State Add all points in $\tilde{Y}^{k-1}$ to $\tilde{Y}^{k}$, and initialize their forces in $z_k$ with the corresponding values in $z_{k-1}$ 
\State Call Oracle to add new points to $\tilde{Y}^{k}$, and initialize their corresponding forces in $z_k$
\State $x_k \gets x_{k-1}$

\LineComment{Solve for step direction}
\State{Set up inner optimization $P^k = P(\tilde{Y}^k)$}
\State Run $S$ steps of an NLP solver on $P^k$, starting from $x_{k},z_{k}$  
\State {Set $x^*,z^*$ to its solution, and $\Delta x \gets x^*-x_{k}$, $\Delta z \gets z^*-z_{k}$}
\State {Do backtracking line search with at most $N_{LS}^{max}$ steps to find optimal step size $\alpha$ such that $\phi(x_{k}+\alpha \Delta x,z_{k}+\alpha \Delta z;\mu) \leq \phi(x_{k},z_{k};\mu)$}

\LineComment{Update state and test for convergence}
\State $x_{k} \gets x_{k} + \alpha \Delta x$, $z_{k} \gets z_{k} + \alpha \Delta z$
\If{Convergence condition is met}
\EndIf
\Return $x_{k}$,$z_{k}$
\EndFor
\State \Return \textsc{Not\;converged}
\end{algorithmic}
\end{algorithm}



\vspace{2mm}
\noindent
\textbf{Infinite programming for contact-rich trajectory optimization.}
First, we describe the IP formulation used by STOCS.  Let us start by defining {\em semi-infinite programming} (SIP).
An SIP problem is an optimization problem in finitely many variables $q \in \mathbb{R}^{n}$ on a feasible set described by infinitely many constraints:
\begin{subequations}
\label{1}
\begin{align}
\min_{q\in \mathbb{R}^n} &\; f(q) \\
\text{s.t.} &\; g(q,y) \geq 0 \quad \forall y \in Y 
\end{align} 
\end{subequations} 
where $g(q,y) \in \mathbb{R}^{m}$ is the constraint function, $y$ denotes the {\em index parameter}, and $Y \in \mathbb{R}^{p}$ is the {\em index domain} which can be continuously infinite. 

As an example, to solve pose optimization with non-penetration constraints~\cite{hauser2021semi}, $q$ describes the pose of the object, and $y$ is a point on the surface of the object $\mathcal{O}$, where $Y\equiv \partial \mathcal{O}$ denotes the surface of $\mathcal{O}$.  The constraint $g(\cdot,\cdot)$ is the signed distance from a point to the environment. Although the inequality $g(q,y) \geq 0$ must be satisfied for all $y$, optimal solutions will be supported by some points, i.e., at the optimal solution $q^\star$, the Karush-Kuhn-Tucker conditions will be established by a finite set of points $y$ such that $g(q^\star,y)=0$.

In trajectory optimization, constraints need to be  enforced in space-time~\cite{hauser2021semi}, so we define a candidate index point $y_t \in \partial \mathcal{O}$ as being indexed by time. We denote $Y_t = \partial \mathcal{O}$ as the index domain at time $t$. 

STOCS then reasons about contact forces at each index point, and since the contact force distribution is a function over the object surface, this introduces infinitely many variables, which turns the problem into an infinite program~\cite{anderson2012infinite}.  Specifically, we define the contact force $z(\cdot): Y_t \rightarrow \mathbb{R}^r$ as an optimization variable.  Our formulation imposes friction constraints, complementarity constraints, and quasi-dynamic/quasi-static force and moment balance on these variables.  

To formulate the force variables and friction cone, we encode normal and tangential forces separately in the vector $z$.  In 2D, $z(y)=[z^N,z^+,z^-]$ is expressed in a reference frame with $z^N$ the component of force normal to the contact surface, and $z^+$, $z^-$ tangent to the contact surface. In 3D, following \cite{posa2014direct}, we use a polyhedral approximation of the friction cone \cite{stewart1996implicit}. We express $z(y)=[z^N,z^{D1},\cdots,z^{Dd}]$ in a reference frame with $z^N$ normal to the contact surface, and $z^{D1}, \cdots, z^{Dd}$ tangent to the contact surface. The convex hull of the unit vectors along the directions of $z^{D1}, \cdots, z^{Dd}$ in $\mathbb{R}^2$ forms the polyhedral approximation.  The force is required to satisfy $z\geq 0$ and $\mu z^N - \sum_{i=D_1}^{D_{d}} z^i \geq 0$ where $\mu$ is the friction coefficient of the given point and the right hand side is the sum of tangential forces. 

In summary, the constraints we impose at time $t$ are as follows:
\begin{enumerate}[label={\bfseries\arabic*.}, ref=\arabic*]
    \item \textbf{State and Control Bounds}: \begin{equation} \label{2} q_t\in\mathcal{Q},  \; v_t\in \mathcal{V}, \; u_t\in\mathcal{U} \end{equation}  
    \item \textbf{Object Dynamics}: \begin{equation}q_t - q_{t+1} + v_{t+1}\Delta t = 0 \label{7d}\end{equation}
    \item  \textbf{Distance Complementarity}: ensures that nonzero forces are only exerted at points where objects are in contact, i.e. the normal force $z^N(y)$ is nonzero only if contact is made at $y$.  \begin{equation} \label{3}  0 \leq z^N(y) \perp g(q_t,y) \geq 0 \quad \forall y \in Y_t \end{equation} 
    \item \textbf{Unilateral Manipulation Velocity}: ensures the manipulator can only push the object rather than pull.
    \begin{equation} \label{5} c(q_t,u_t) \geq 0 \end{equation} 
    \item \textbf{Force-torque Balance}: \begin{equation} \label{6} \underbrace{ s_{q,u}(q_t,u_t) + \int_{y \in Y_t} s_z(q_t,y,z(y)) dy}_{\eqqcolon s(q_t,u_t,z;Y_t)} = M \dot{v}_t \;\text{or} \;0\end{equation} 
    \item \textbf{Friction constraints}: the force constraints defined above are grouped into a function $h$ \begin{equation}
    h(q_t,y,z(y)) \geq 0 \quad  \forall y\in Y_t.
    \label{eq:friction}
    \end{equation}
    A similar constraint is applied to the manipulator contact force.
    \item \textbf{Friction-Velocity Complementarity}: ensures the relative tangential velocity at a contact is zero unless the contact force is at the boundary of the friction cone
    \begin{equation} 0 \leq w(q_t,v_t,y) \perp  h(q_t,y,z(y)) \geq 0 \quad \forall y\in Y_t.
    \end{equation} 
\end{enumerate}
We elaborate a bit on~\eqref{6}, which integrates the force distribution over the domain $Y_t$. Here, $s_{q,u}(q,u)$ represents the force and torque applied by gravity and the manipulator, and $s_z(q,y,z(y))$ represents the force and torque applied on an index point $y$ by the contact force $z(y)$. The integral gives the net force and torque experienced by the object, which is $0$ if we assume quasi-static, and is $M\dot{v}_t$ if we assume quasi-dynamic, where $M$ is the inertia matrix of $\mathcal{O}$, and $\dot{v}_t$ is approximated as $v_t/\Delta t$.

Lastly, adding terminal constraints $q_0 \in Q_{init}$ and $q_T \in Q_{goal}$, we formulate the following infinite programming with complementarity constraints trajectory optimization (IPCC-TO) problem:
\begin{subequations}
\label{pb:3}
\begin{align}
\min_{q,v,u,z} &\; 
f(q,v,u,z)\\
\text{s.t.} & \; q_0\in \mathcal{Q}_{init}, q_T\in \mathcal{Q}_{goal} \label{7b}\\ 
&\; (2),(4),(5),(6),(8) \quad t = 0,\ldots,T \label{7c}\\
&\; (3) \quad t = 0,\ldots,T-1, \label{7d}
\end{align}
\end{subequations}
where $q$, $v$, and $u$ concatenate the state and control variables along all time steps, and $z$ represents the contact force distribution at all points in time. Detailed instantiations of Equation \ref{pb:3} for both 2D and 3D scenarios are available in the Appendix. We denote this problem as $P(Y)$.

\vspace{2mm}
\noindent
\textbf{Exchange method.}
The IPCC-TO problem $P(Y)$ not only has infinitely many constraints, but also an infinity of variables in $z$. To solve it using numerical methods, we require that $z$ only is non-zero at a finite number of points. Indeed, if an optimal solution $q^\star$ is supported by a finite subset of index points $\tilde{Y} \in Y$, then it suffices to solve for the values of $z$ at these supporting points, since $z$ should elsewhere be zero thanks to \eqref{3}. This concept is used in the exchange method to solve SIP problems~\cite{lopez2007semi}, and we extend it to solve IPCC-TO. 

The exchange method progressively instantiates finite index sets $\tilde{Y} \equiv [\tilde{Y}_0, \cdots, \tilde{Y}_T]$ and their corresponding finite-dimensional MPCCs whose solutions converge toward the true optimum~\cite{reemtsen1998numerical}.  The solving process can be viewed as a bi-level optimization. In the outer loop, index points are selected by an oracle to be added to the index set $\tilde{Y}$, and then in the inner loop, the optimization $P(\tilde{Y})$ is solved. The outer loop will then decide how much should move toward the solution of $P(\tilde{Y})$. Specifically, if we let $(\tilde{x}^*=[q^*,v^*,u^*],\tilde{z}^*)$ be the optimal solution to $P(\tilde{Y})$, then as $\tilde{Y}$ grows denser, the iterates of ($\tilde{x}^*$, $\tilde{z}^*$) will eventually approach an optimum of $P(Y)$. 

Given a finite number of instantiated contact points $\tilde{Y} \subset Y$, we can solve a discretized version of the problem which only creates constraints and variables corresponding to $\tilde{Y}$. Also, through replacing the distribution of $z(y)$ with Dirac impulses, integrals are replaced with sums and we formulate the finite MPCC problem $P(\tilde{Y})$ in the following form:
\begin{subequations}
\begin{align}
\label{pb:4}
\min_{q,v,u,z} &\; \tilde{f}(q,v,u,z) \\
\text{s.t.}& \; q_0\in \mathcal{Q}_{init}, q_T\in \mathcal{Q}_{goal}\\
&\; (2),(4),(5),(8) \quad t = 0,\ldots,T\\
&\; (3) \quad t = 0,\ldots,T-1\\
&\; \tilde{s}(q_t,u_t,z;\tilde{Y}_t)=M \dot{v} \; or\; 0 \quad t=0,\cdots,T
\end{align}
\end{subequations}
where $\tilde{f}(q,v,u,z) \coloneqq \sum_{t=0}^{T} [f_{q,v,u}(q_t,v_t,u_t) + \sum_{y \in \tilde{Y}_t} f_z(q_t,y,z(y))] $, and $\tilde{s}(q_t,u_t,z;\tilde{Y}_t) = s_{q,u}(q_t,u_t) + \sum_{y \in \tilde{Y}_t} s_z(q_t,y,z(y))$. 

\vspace{5pt}
\noindent
\textbf{Oracle.}
Identifying an effective selection strategy for index points is a key to solving the IPs, and this strategy is denoted the Oracle. A naive Oracle would sample index points incrementally from $Y_t$ (e.g., randomly or on a grid) at each of the discretized time steps along the trajectory, and hopefully, with a sufficiently dense set of points the iterates of solutions will eventually approach an optimum. However, this approach is inefficient, as most new index points will not yield active contact forces during the iteration.  Better strategies seek to identify a small number of points that would be active at an optimal solution.  This work introduces two different Oracle designs, the Maximum Violation Oracle (MVO) and the Time-active Maximum Violation Oracle (TAMVO), which will be discussed in more detail in Sec.~\ref{TAMVO}.

\noindent
\textbf{Merit function for the outer iteration.}
After solving $P(\tilde{Y})$ in an outer iteration, we get a step direction from the current iterate ($\tilde{x}$, $\tilde{z}$) toward ($\tilde{x}^*$, $\tilde{z}^*$), where $x=[q, v, u]$ is a concatenation of all the optimization variables except for $z$. However, due to nonlinearity, the full step may lead to a worse constraint violation for the original problem $P(Y)$. To avoid this, we perform a line search over the following merit function that balances reducing the objective and reducing the constraint error on the infinite dimensional problem $P(Y)$:

\begin{equation}
\phi(x, z; \mu) = f(x,z) + \mu \|b(x,z)\|_1,
\end{equation}
where $b$ denotes the vector of constraint violations of Problem (9). Also, in SIP for collision geometries, a serious problem is that using existing instantiated index parameters, a step may go too far into areas where the minimum of the distance function $g^*(q_t)\equiv \min_{y\in Y_t} g(q_t,y)$ greatly violates the inequality, and the optimization loses reliability. So we add the max-violation $g^{*-}(q_t)$ to $b$, in which we denote the negative component of a term as $\cdot^- \equiv \min(\cdot,0)$. 

\vspace{2mm}
\noindent
\textbf{Convergence criteria.}
We denote the index set $\tilde{Y}$ instantiated at the $k^{th}$ outer iteration as $\tilde{Y}^{k}$, the corresponding MPCC as $P_k=P(\tilde{Y}^k)$, and the solved solution as ($x_k$, $z_k$).

The convergence condition is defined as $\alpha \|[\Delta x,\Delta z]\| \leq \epsilon_x\cdot n_{xz}$ \text{and}
$|z_{k}|^T |g(q_{k},\tilde{Y}^{k})| + |v(q_{k},v_k,\tilde{Y}^k)|^T |h(q_k,\tilde{Y}^k,z_k)|\leq \epsilon_{gap} \cdot n_{cc}$ \text{and}
$|s(x_{k},z_{k},\tilde{Y}^{k})| \leq \epsilon_s \cdot T$ (or \quad 
$|s(x_{k},z_{k},\tilde{Y}^{k})| - M \dot{V} \leq \epsilon_s \cdot T$) \text{and}
$\sum_t g_{k}^{-*}(x_{k,t}) < \epsilon_{p}\cdot T$, where $n_{xz}$ is the dimension of the optimization variable and $n_{cc}$ is the number of complementarity constraints, $\epsilon_x$ is the step size tolerance, $\epsilon_{gap}$ is the complementarity gap tolerance, $\epsilon_{s}$ is the balance tolerance, and $\epsilon_{p}$ is the penetration tolerance. With a little abuse of notation, $g(x_{k},\tilde{Y}^{k})$ is the concatenation of the function value of all the points in $\tilde{Y}^k$, and similar for $v(q_{k},v_k,\tilde{Y}^k)$ and $h(q_k,\tilde{Y}^k,z_k)$. $\dot{V}$ is the concatenation of all the $\dot{v}_t$ for $t \in \mathcal{T}$.

\subsection{Oracle Choice} \label{TAMVO}

As described above, the oracle design is a key component of STOCS. We compare Maximum Violation Oracle (MVO), that adds the closest / deepest penetrating points between the object and the environment at each time step along the trajectory, along with a new method, Time Active Maximum Violation Oracle (TAMVO) with smoothing.
TAMVO selects index points more judiciously and only adds the closest / deepest penetrating points at a specific time step.

\begin{algorithm}[]

\caption{Maximum-Violation Oracle}
\label{alg:3}
\begin{algorithmic}[1]
\Statex \textbf{Input} $q_{0:T}$, $\tilde{Y}^{k-1}$, adding threshold $d_{min}^*$ and $d_{max}^*$
\Statex \textbf{Output} $\tilde{Y}_{k}$
\State $\tilde{Y}^{k} \gets \tilde{Y}^{k-1}$
\For{$t=0,\ldots,T$}
    \State $y^*=\argmin_{y \in Y_t} g(q_{t},y)$
    \State $d^*=g(q_{t},y^*)$
    \If{$y^*$ is not in $\tilde{Y}^{k}_t$ and $d^*<d_{max}^*$}
        \If{$\|y-y^*\|>d_{min}^* \; \forall y \in \tilde{Y}^{k}_t$}
        \For {$t=0,\ldots,T$}
            \State add $y*$ to $\tilde{Y}^{k}_t$
        \EndFor
        \EndIf
    \EndIf
\EndFor

\end{algorithmic}
\end{algorithm}

\begin{figure}
    \centering
    \includegraphics[width=1\linewidth]{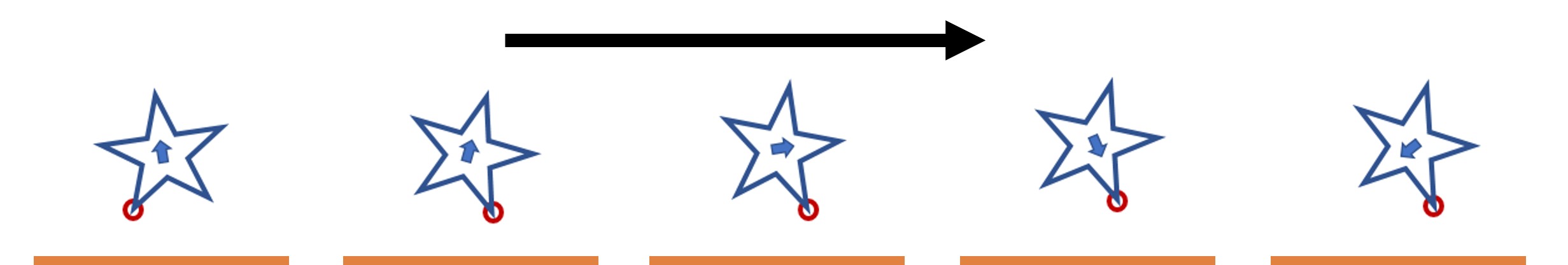}
    \makebox[1.0\linewidth]{\footnotesize (a) Closest point on the object to the environment at each time step}\\
    \includegraphics[width=1\linewidth]{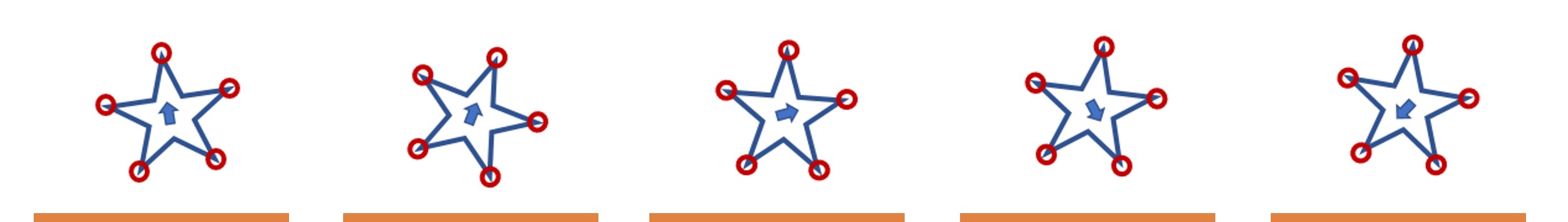}\\
    \makebox[1.0\linewidth]{\footnotesize (b) Index points selected by MVO at each time step}\\
    \includegraphics[width=1\linewidth]{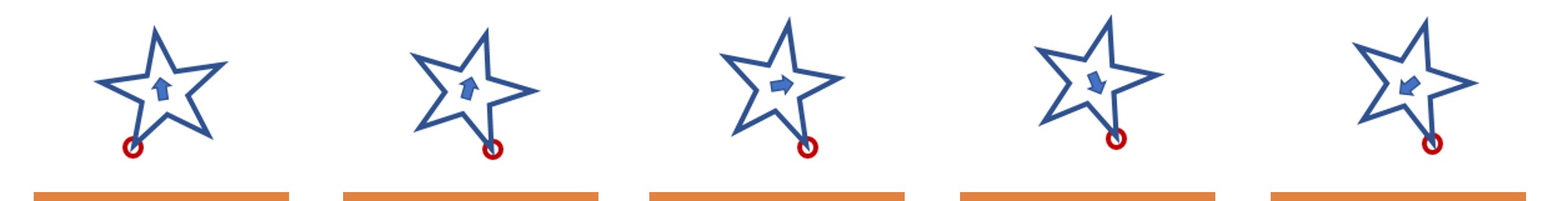}\\
    \makebox[1.0\linewidth]{\footnotesize (c) Index points selected by TAMVO without SD and TS at each time step}\\
    \includegraphics[width=1\linewidth]{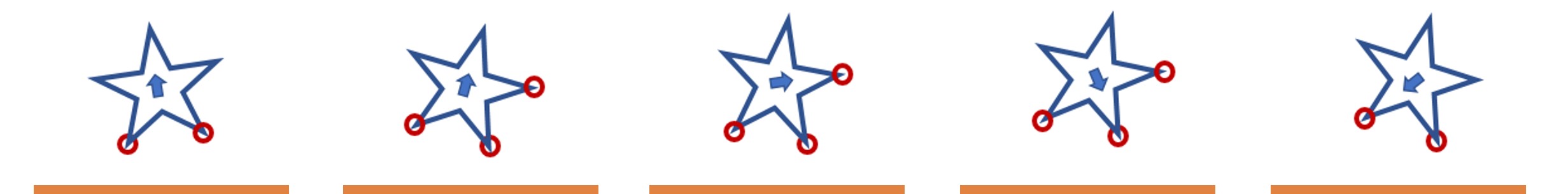}\\
    \makebox[1.0\linewidth]{\footnotesize (d) Index points selected by TAMVO with TS ($n_s=1$) at each time step} 
    \includegraphics[width=0.25\linewidth]{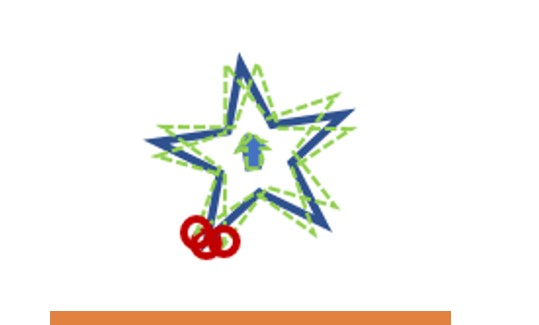}\\
    \makebox[1.0\linewidth]{\footnotesize (e) Spatial Disturbance}
    \caption{Comparing various Oracles. The object trajectory is depicted as moving from left to right (as indicated by the black arrow) and undergoing clockwise rotation (as indicated by the arrow on the star). (b) In Maximum Violation Oracle (MVO), the closest point on the object is added ot the candidate set at every time step. (c) The Time-Active Maximum Violation Oracle, without Spatial Disturbance and Spatial Disturbances, introduces the closest point only at the current time step. The Time Smoothing technique with $n_s=1$, demonstrated in (d), extends constraint imposition to the closest points identified at adjacent time steps. (e) presents the Spatial Disturbance technique applied at a specific time step, with only disturbed rotation illustrated. [Best viewed in color.]}
    \label{fig:oracle}
\end{figure}

MVO (Alg.~\ref{alg:3}) adds the closest or deepest penetrating points between the object and environment at each time step along the trajectory to all candidate index points across time $\tilde{Y}_t^k$s.  Note, however, that it may still include index points that do not generate active contact forces during the iteration. For instance, as illustrated in Fig.~\ref{fig:oracle}(a), the closest or deepest penetrating points at time step $t$ are typically active only around that specific period.

\begin{algorithm}[]

\caption{Time-Active Maximum-Violation Oracle}
\label{alg:4}
\begin{algorithmic}[1]
\Statex \textbf{Input} $q_{0:T}$, $\tilde{Y}^{k-1}$, max object-environment distance $d_{max}^*$, contact uniqueness threshold $\epsilon$, time smoothing step $n_t$, spatial disturbances $N_s$
\Statex \textbf{Output} $\tilde{Y}^{k}$
\State $\tilde{Y}^{k} \gets \tilde{Y}^{k-1}$
\State $\tilde{Y}' \gets [[\;]_{0}, [\;]_{1}, \ldots, [\;]_{T}]$
\For{$t=0,\ldots,T$}
    \State $y^{*}=\argmin_{y \in \tilde{Y}_t} g(q_{t},y)$
    \State $d^{*}=g(q_{t},y^{*})$
    \If{$d^{*}<d_{max}^*$}
        \State add $y^{*}$ to $\tilde{Y}'[t]$
    \EndIf
\EndFor                
\For{$t=0,\ldots,T$}
    \For{$n_s \in N_s$}
        \State $y_s=\argmin_{y \in Y_t} g(q_{t} + n_s,y)$
        \If{$g(q_{t} + n_s,y_s) < d_{max}^*$}
        \State add $y_s$ to $\tilde{Y}'[t]$
        \EndIf
    \EndFor
\EndFor
\For{$t=0,\ldots,T$}
    \For{$t'=t-n_t,\ldots,t+n_t$}
        \If{$0\leq t' \leq T$}
            \For{$y'$ in $\tilde{Y}'[t]$}
                 \If{$\|y'-y\|>\epsilon \; \forall y \in \tilde{Y}^{k}_t$}
                 \State add $y'$ to $\tilde{Y}^{k}_t$
                 \EndIf
            \EndFor
        \EndIf
    \EndFor
\EndFor
\end{algorithmic}
\end{algorithm}

To address this issue, we introduce TAMVO (Alg.~\ref{alg:4}).  In this refined approach, the index set is no longer the same across time steps.  Lines 8--12 identify closest points at each time step.  Duplicate points (within threshold $\epsilon$) are excluded in lines 13--18.   Given default parameters $n_t=0$ and $N_s=[0]$, adds only the closest or most deeply penetrating points at the current time step $t$ to $\tilde{Y}_t^k$. 

Choosing only the closest points at the current iterate is potentially not the most ideal choice unless the current iterate is near-optimal. A better choice would anticipate which points are active at the optimum. To address this, we introduce the following two strategies designed to mitigate this issue.

\noindent
\textbf{Spatial Disturbance (SD).}
Recognizing that the current iterate is likely to be in the neighborhood of the optimal solution, the SD approach introduces perturbations to the current solution to add new candidate contact points. Consequently, in lines 9--12 of Alg.~\ref{alg:4}, $q_t$ is perturbed with disturbance $n_s$ to choose closest points. We choose to perturb along each dimension of $q_t$ in both positive and negative directions. In 3D, this strategy chooses 12 perturbations accounting for both increases and decreases in $x$, $y$, $z$ and $roll$, $pitch$, $yaw$.  An illustration of adding perturbation to rotation in 2D is shown in Fig.~\ref{fig:oracle}(e).  

\noindent
\textbf{Time Smoothing (TS).}
Considering that the closest points may be active not just at the current time step $t$, but also during a surrounding interval, in line 14 of Alg.~\ref{alg:4}, the closest points detected within the adjacent time steps from $t-n_t$ to $t+n_t$, governed by a parameter $n_t$, are added to the index set of time step $t$. The effect of using TS is illustrated in Fig.~\ref{fig:oracle}(d).

\section{Experimental Results}\label{sec:experiment}

The proposed methods are implemented in Python using the optimization interface and the SNOPT solver \cite{gill2005snopt} provided by Drake \cite{drake}.

\subsection{Experiments in 2D}
First, we compare STOCS with vanilla MPCC to evaluate the efficacy of dynamic contact selection on an object pivoting task. We finely discretize the object geometries to better illustrate the advantages of our method. Vanilla MPCC involves adding all index points in $Y$ to an MPCC problem without selection, resulting in a larger optimization problem than $P^k$ in STOCS. MVO is employed in this experiment, and $T=20$, $\Delta T=0.1\;s$, $\mu_{mnp}=1.0$ and $\mu_{env}=0.5$ are used for this set of the experiments.

\begin{figure}[tbp]
    \centering
    \includegraphics[width=0.98\linewidth]{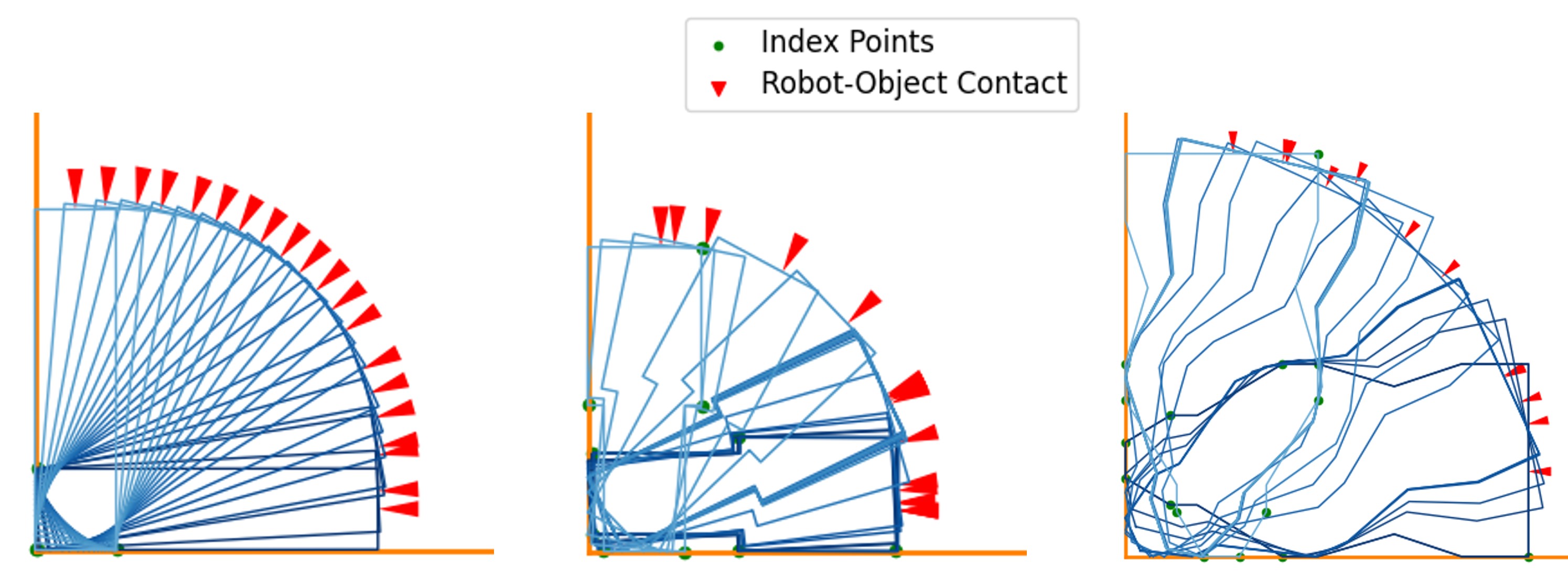}
    \caption{Pivoting trajectories for 3 objects solved by STOCS.  The object's poses, the contact between the object and the robot (red triangle), and object-environment contact points (green dots) are plotted for each time step. [Best viewed in color.]}
    \label{fig:2D_pivot}
\end{figure}

The results are presented in Table~\ref{tab:pivot}. We observe that STOCS can be around two to three orders of magnitude faster than MPCC, and can solve problems that MPCC cannot solve. STOCS selects only a small amount of points from the total number of points in the objects’ representation on average, which greatly decreases the dimension of the instantiated optimization problem and reduces solve time. The solved trajectories of STOCS for three different object are shown in \ref{fig:2D_pivot}.

\begin{table}[tbp]
\caption{Numerical results on the pivoting task. Number of points in the object's representation (\# Point), solve time (Time), outer loop iteration number (Outer iters), and average active index points for each iteration (Index points) are reported in the table. \label{tab:pivot}}
\centering
\renewcommand{\arraystretch}{1.2}
\setlength\tabcolsep{4pt}
\begin{tabular}{@{}llllll@{}}
\toprule
\multicolumn{2}{c}{}                                            & \multicolumn{3}{c}{{\bf STOCS}}                                                                                                                                 & {\bf MPCC}     \\ \cline{3-5}
Object  & \# Points &  Time (s) & Outer iters. & Index points & Time (s) \\ \midrule
Box     &   212    &    4.3    &     2                                                      &   2.0   &    2908.5    \\   
Peg     &   214    &    45.2    &  5    &  4.2   &    5503.9      \\ 
Mustard &   400 &   33.6   &   5   &  6.8     &  Failed     \\ \bottomrule
\end{tabular}
\end{table}

\begin{figure}[tbp]
    \centering
    \includegraphics[width=0.8\linewidth]{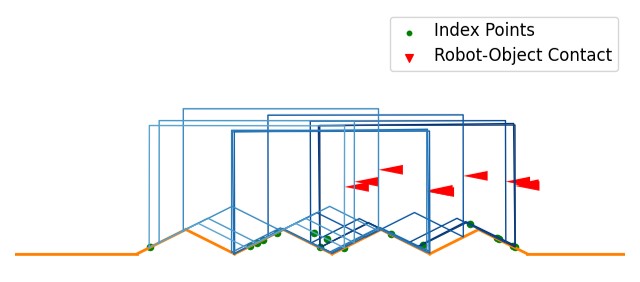}
    \makebox[1.0\linewidth]{\footnotesize (a) Pushing a dented object on uneven terrain}\\
    \includegraphics[width=0.8\linewidth]{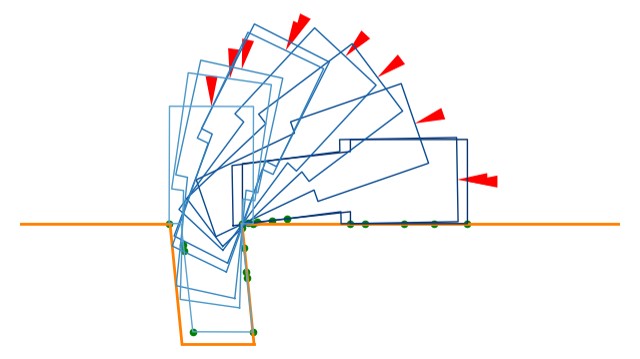}\\
    \makebox[1.0\linewidth]{\footnotesize (b) Pushing a tilted peg into angled hole}\\
    \includegraphics[width=0.8\linewidth]{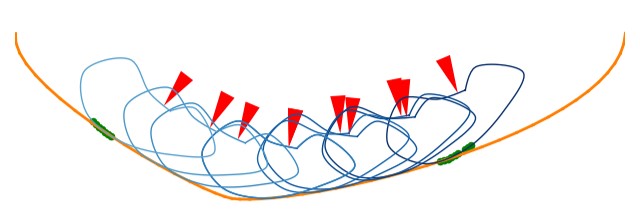}\\
    \makebox[1.0\linewidth]{\footnotesize (c) Sliding bean on curve 1}\\
    \includegraphics[width=0.8\linewidth]{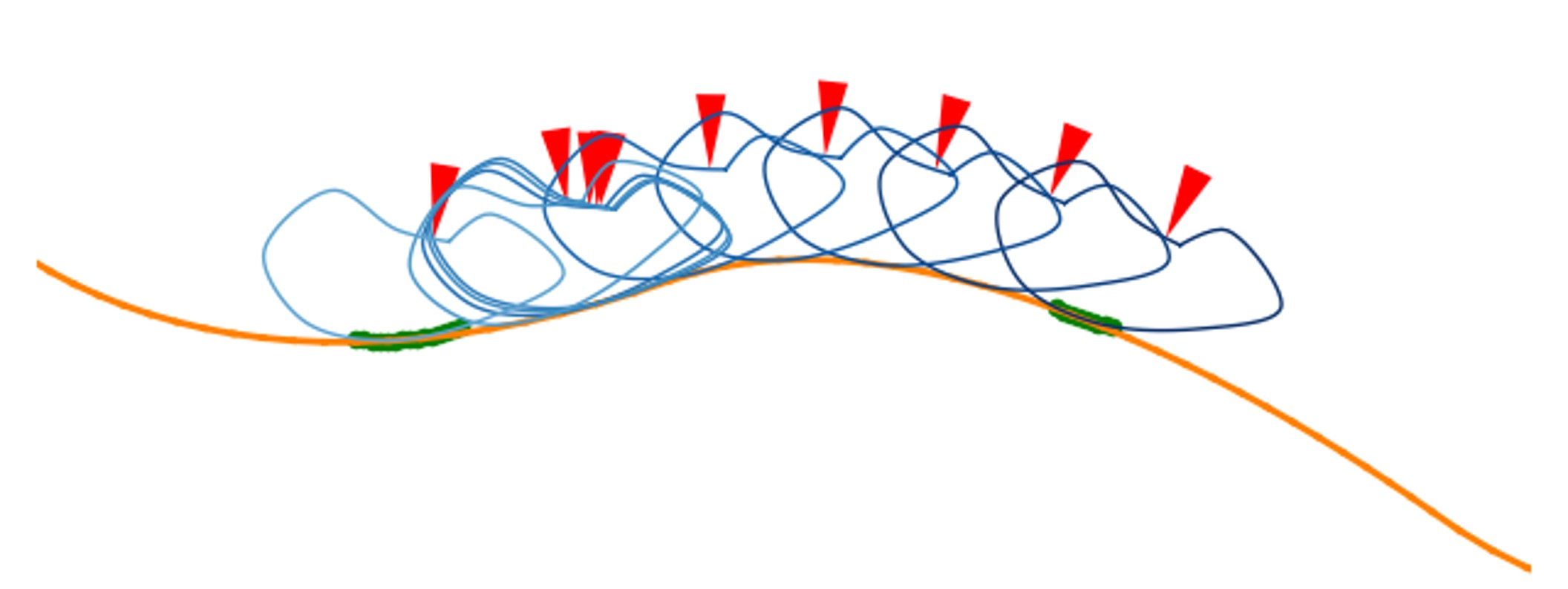}\\
    \makebox[1.0\linewidth]{\footnotesize (d) Sliding bean on curve 2}\\
    \caption{Trajectories planned by STOCS on 2D examples. The progression from the start to the end of the trajectory is indicated by a dark to light gradient. For clarity, the instantiated object-environment contact points are depicted only at the first and the last time steps. [Best viewed in color.]}
    \label{fig:2D_trajectories}
\end{figure}

Next, we evaluate STOCS in a 2D setting with TAMVO, focusing on testing its applicability with objects and environments whose geometries are significantly more complex than those presented in the examples of \cite{zhang2023simultaneous}. In this experiment, we set $n_t=1$ and $N_s=[1e^{-2}]$ as the parameters for TAMVO, and we designate these as the default values for TS and SD separately. Four tasks are designed for this evaluation: pushing a dented object across uneven terrain, inserting a tilted peg into a correspondingly angled hole, and maneuvering a bean-shaped object across two distinct curvilinear terrains. The trajectories planned for these tasks are depicted in Fig.~\ref{fig:2D_trajectories}, and the detailed information regarding the objects’ geometries and the solve of the trajectories are presented in Table~\ref{tab:2D}.

$T=10$, $\Delta T=0.1\;s$, $\mu_{mnp}=1.0$ and $\mu_{env}=0.5$ are used for this set of the experiments in 2D.

Same as before, STOCS selects only a small number of points from the entire set in the object’s representation to establish a solution. Typically, STOCS reaches convergence after just a few iterations. The final example, sliding the bean along curve 2, demands additional iterations and time for convergence due to the curve’s curvature shifting from concave to convex. This alteration results in different contact points at different stages along the trajectory, which takes more iterations to be fully instantiated.

\begin{table}[bp]
\caption{Numerical results of STOCS in 2D. Number of points in the object's representation (\# Point), solve time (Time), outer iteration count (Outer iters), and average active index points for each iteration (Index points) are reported.\label{tab:2D}}
\centering
\renewcommand{\arraystretch}{1.2}
\setlength\tabcolsep{4pt}
\begin{tabular}{@{}llllll@{}}
\toprule
Environment & Object                & \# Point             & Outer iters. & Index points & Time (s) \\ \midrule
Uneven      & Dented                & 543                  & 4            & 9.05        & 30.83    \\
Tilted Hole & Tilted Peg            & 214                  & 4            & 12.93        & 40.11     \\
Curve 1     & \multirow{2}{*}{Bean} & \multirow{2}{*}{100} & 7            & 9.43         & 72.74     \\
Curve 2     &                       &                      & 14           & 14.29        & 636.89    \\ \bottomrule
\end{tabular}
\end{table}

\subsection{Experiments in 3D}
\begin{figure*}
    \centering
    \vspace{-5pt}
    \includegraphics[width=0.95\linewidth]{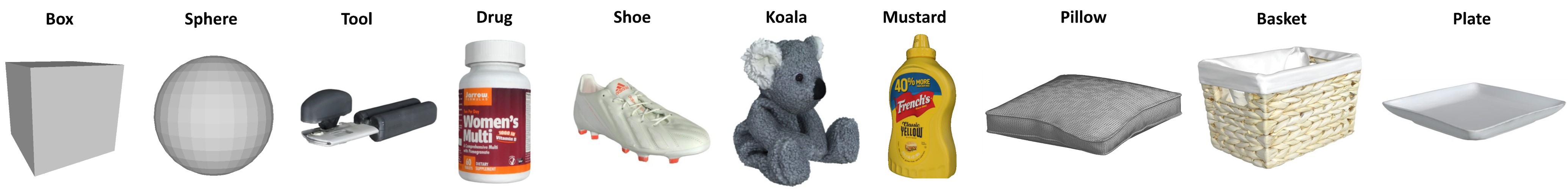}
    \makebox[1.0\linewidth]{\footnotesize (a) Objects}\\
    \vspace{5pt}
    \includegraphics[width=0.95\linewidth]{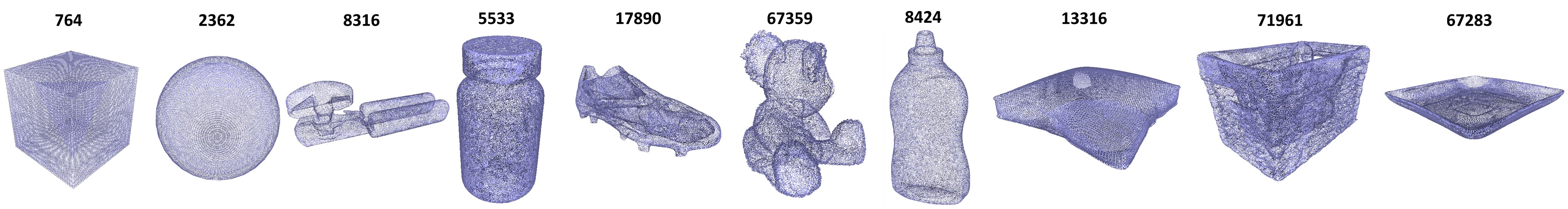}\\
    \vspace{-5pt}
    \makebox[1.0\linewidth]{\footnotesize (b) Point Clouds}\\
    \caption{3D test objects (first row) and their point clouds (second row) used in the experiments. Not drawn to scale.}
    \label{fig:objects}
\end{figure*}

\begin{figure}
    \centering
    \includegraphics[width=1.0\linewidth]{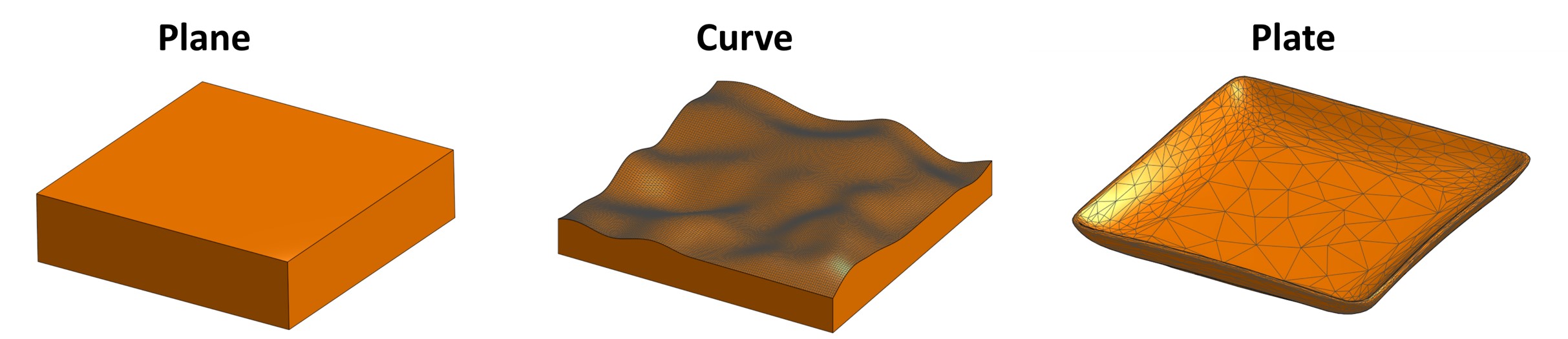}\\
    \vspace{5pt}
    \includegraphics[width=0.7\linewidth]{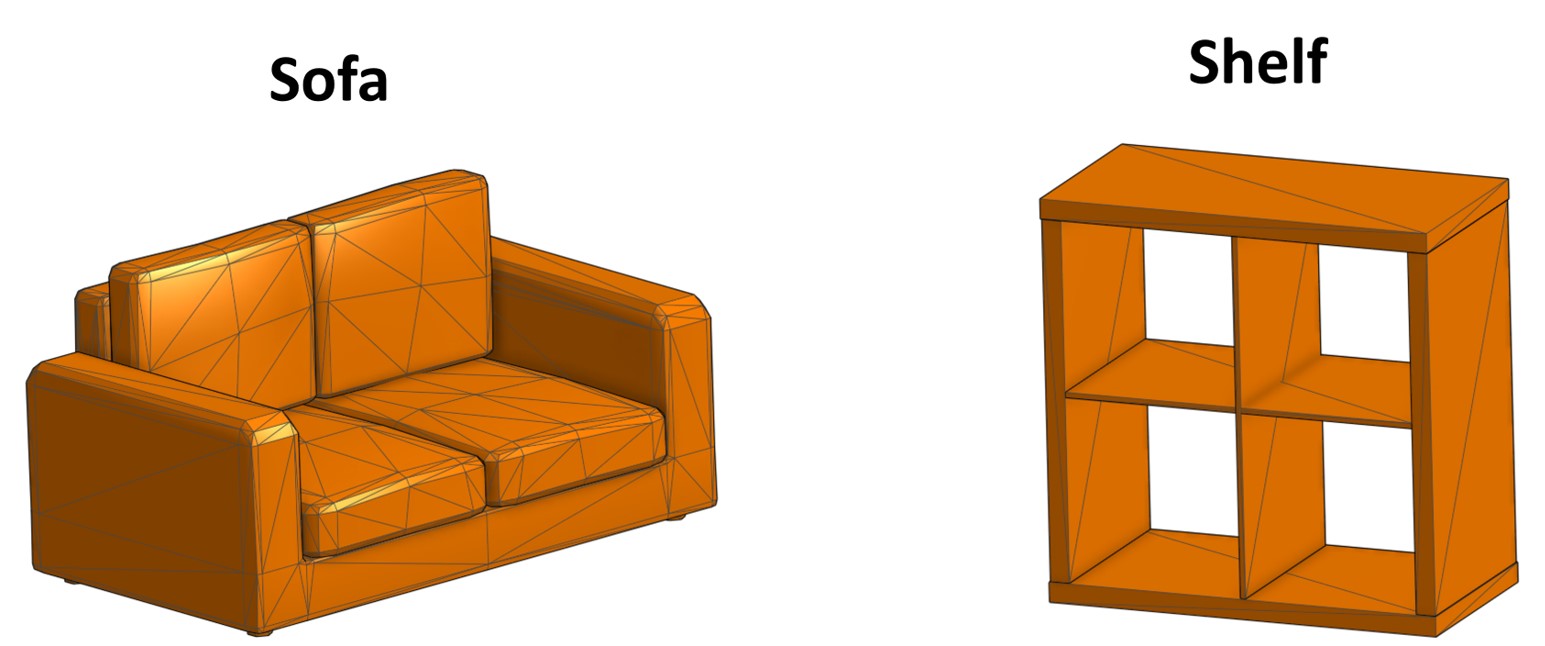}\\
    \caption{3D test environments.  Not drawn to scale.}
    \label{fig:environments}
\end{figure}

Next, we evaluate STOCS in 3D. To demonstrate its generalizability, we collect object geometries from the YCB dataset \cite{calli2015benchmarking}, the Google Scanned Objects \cite{downs2022google}, and 3D models found online \cite{GazeboFuel-OpenRobotics-Sofa,Sketchfab-Pillow}. All the objects used in the study are shown in Fig.~\ref{fig:objects}, and all the environments used in the study are shown in Fig.~\ref{fig:environments}. Klampt \cite{klampt} and the code in \cite{hauser2021semi} are used to find the closest points between two complex shaped geometries. 


\begin{figure}[h]
    \centering
    \includegraphics[width=0.9\linewidth]{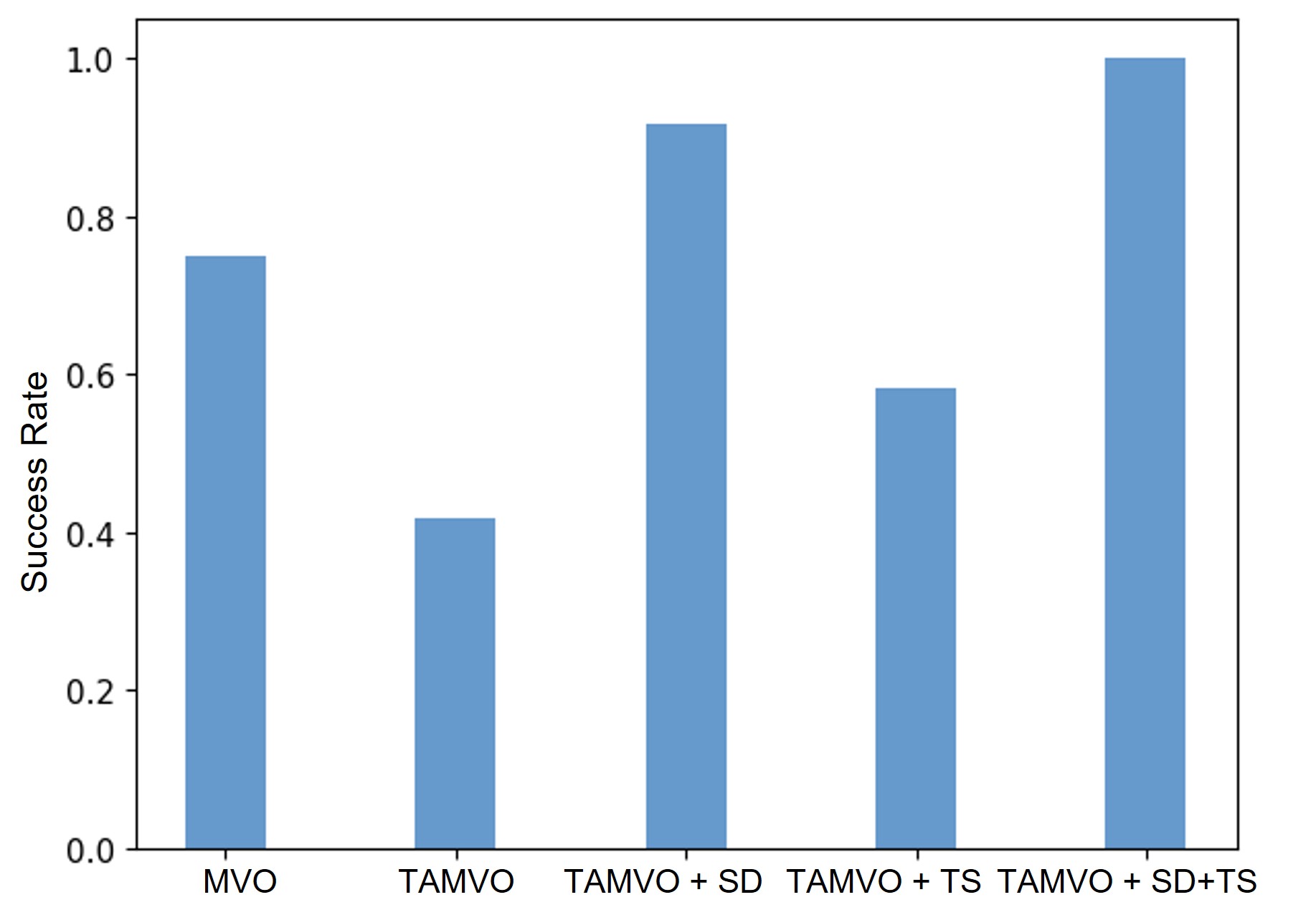}
    \caption{Success rates of STOCS for varying choice of Oracle.} 
    \label{fig:sf}
\end{figure}

\begin{figure}
    \centering
    \includegraphics[width=0.95\linewidth]{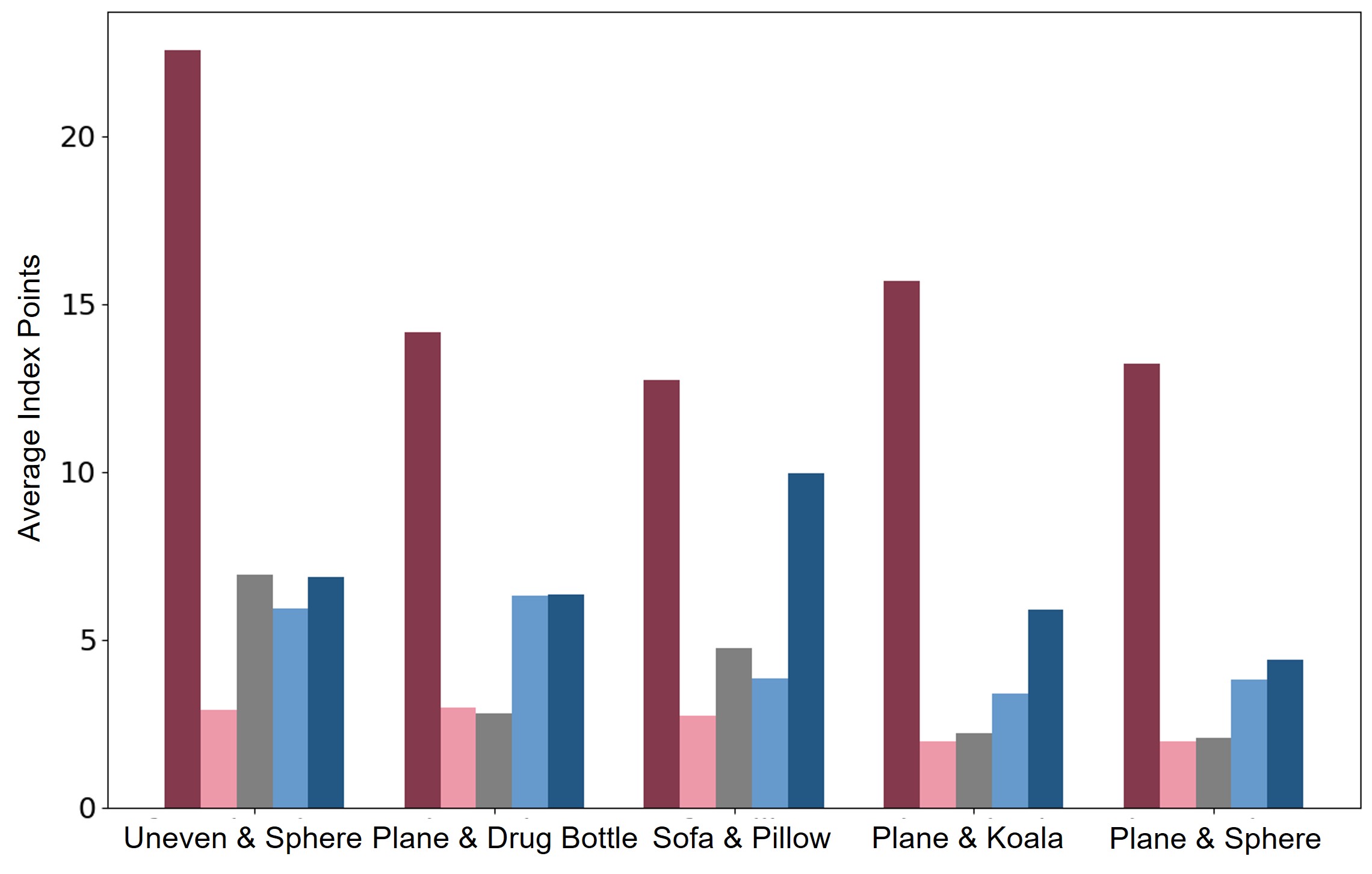}\\
    \makebox[1.0\linewidth]{\footnotesize (a) Average Index Points}\\
    \vspace{5pt}
    \includegraphics[width=0.95\linewidth]{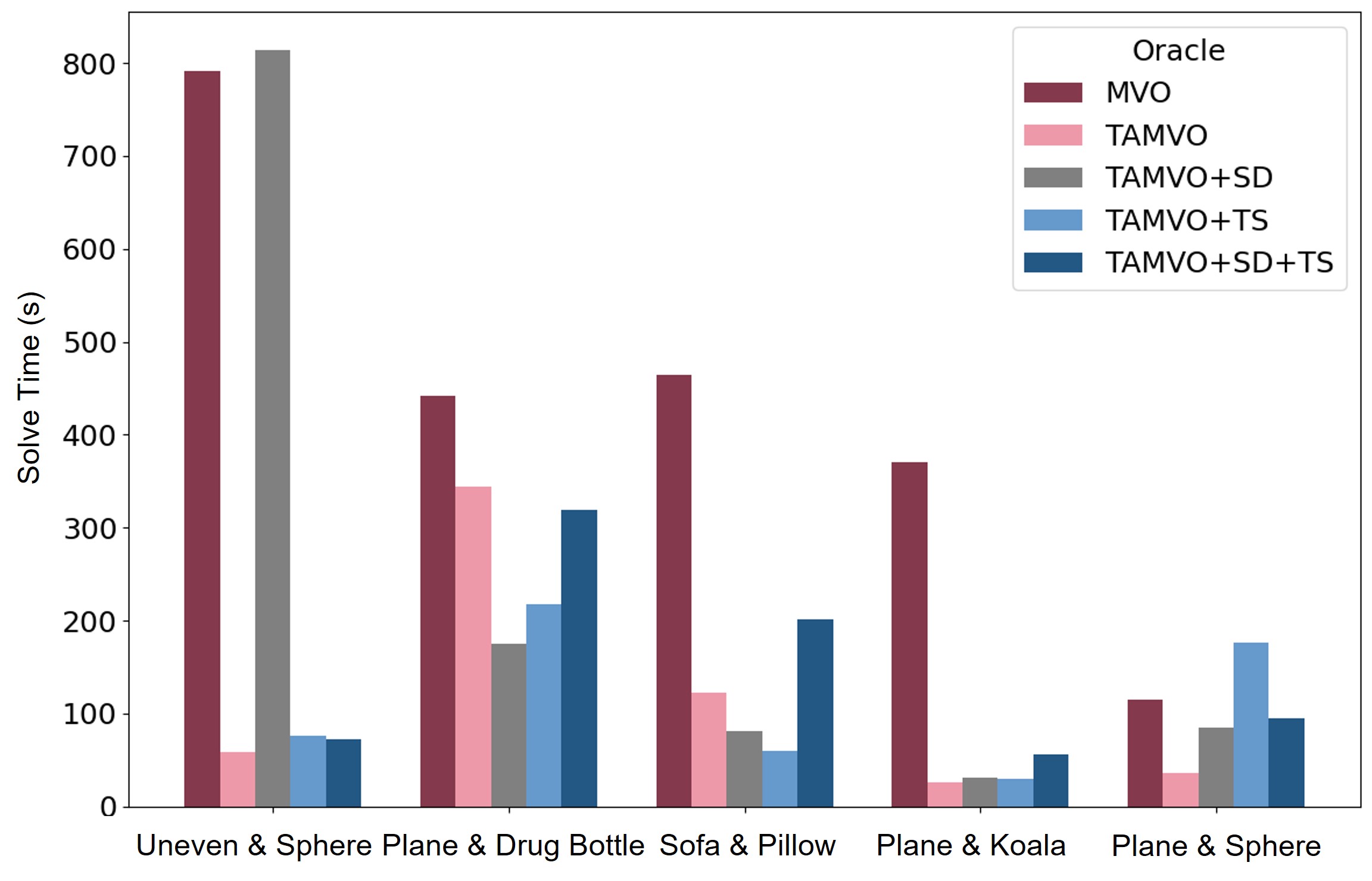}\\
    \makebox[1.0\linewidth]{\footnotesize (b) Solve Time}\\
    \caption{The average index points selected at each time step by the different Oracles (a) and the solve time (b) for tasks that were successfully solved by all Oracles. [Best viewed in color.]} 
    \label{fig:time_and_index}
\end{figure}

As demonstrated in the first set of experiments, employing a vanilla MPCC approach by incorporating all index points in 
$Y$ into an MPCC problem without selection results in a large optimization problem. Such an approach can take a long time to find a solution. Consequently, we have opted not to pursue experiments with the vanilla MPCC in this study, given that the 3D point clouds utilized herein contain thousands of points on average. Additionally, approximating the friction cone with a polyhedral model in a 3D context further increases the number of constraints for each index point, exacerbating the complexity beyond that encountered in 2D scenarios.

\begin{figure*}
    \centering
    \begin{tabular}{ccc}
         \includegraphics[width=0.3\linewidth]{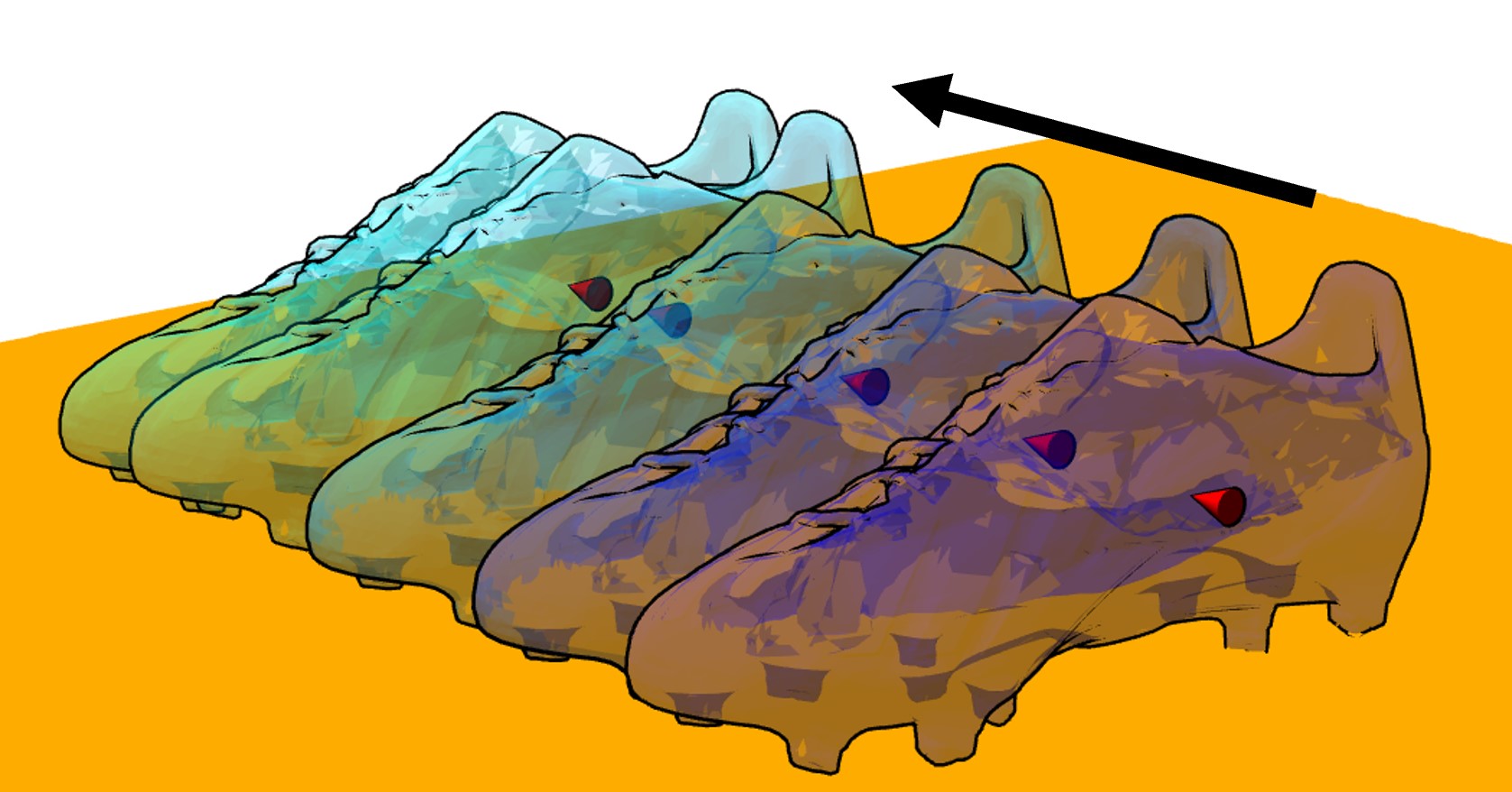}
         &  \includegraphics[width=0.3\linewidth]{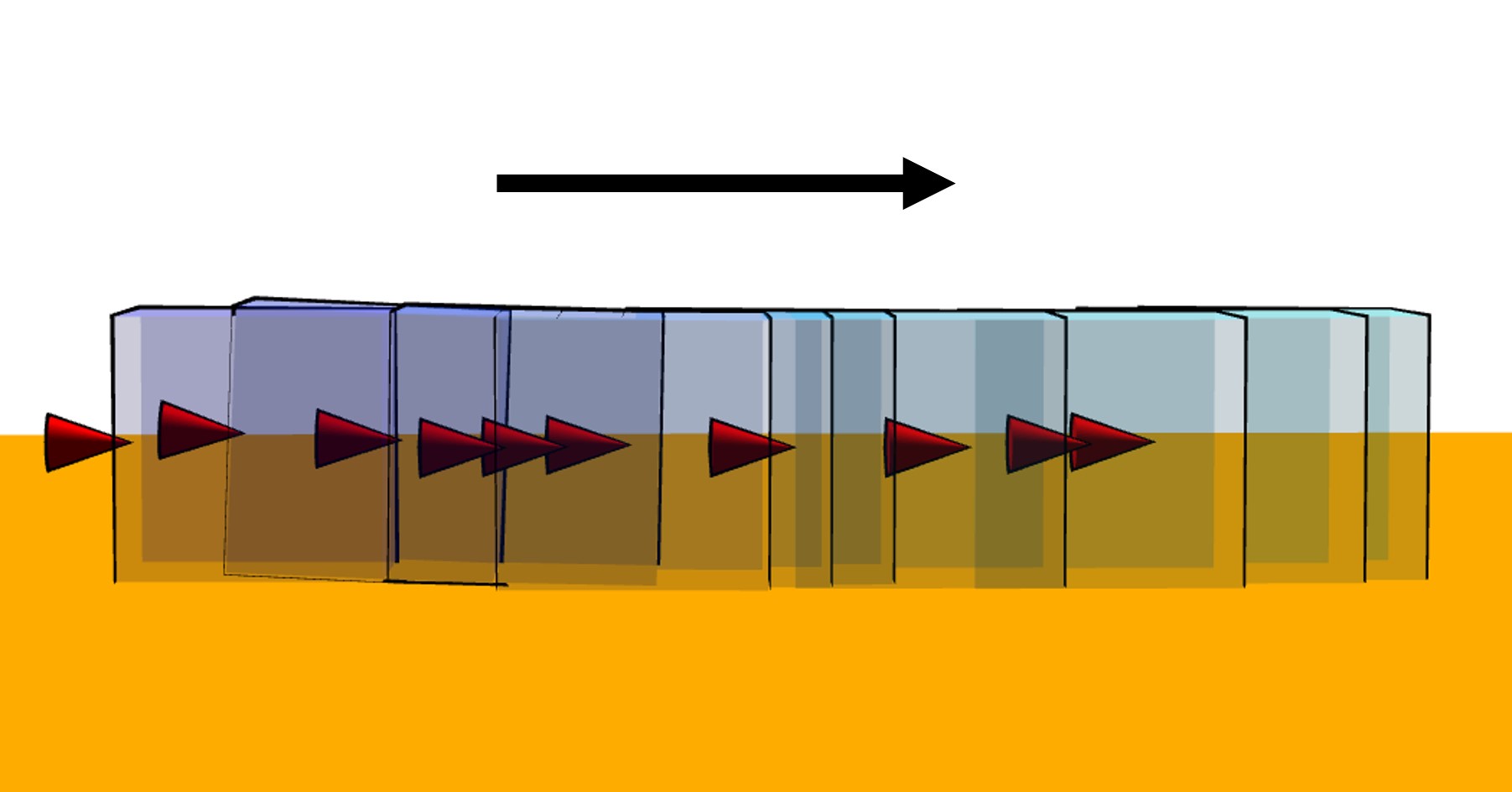} 
         &  \includegraphics[width=0.3\linewidth]{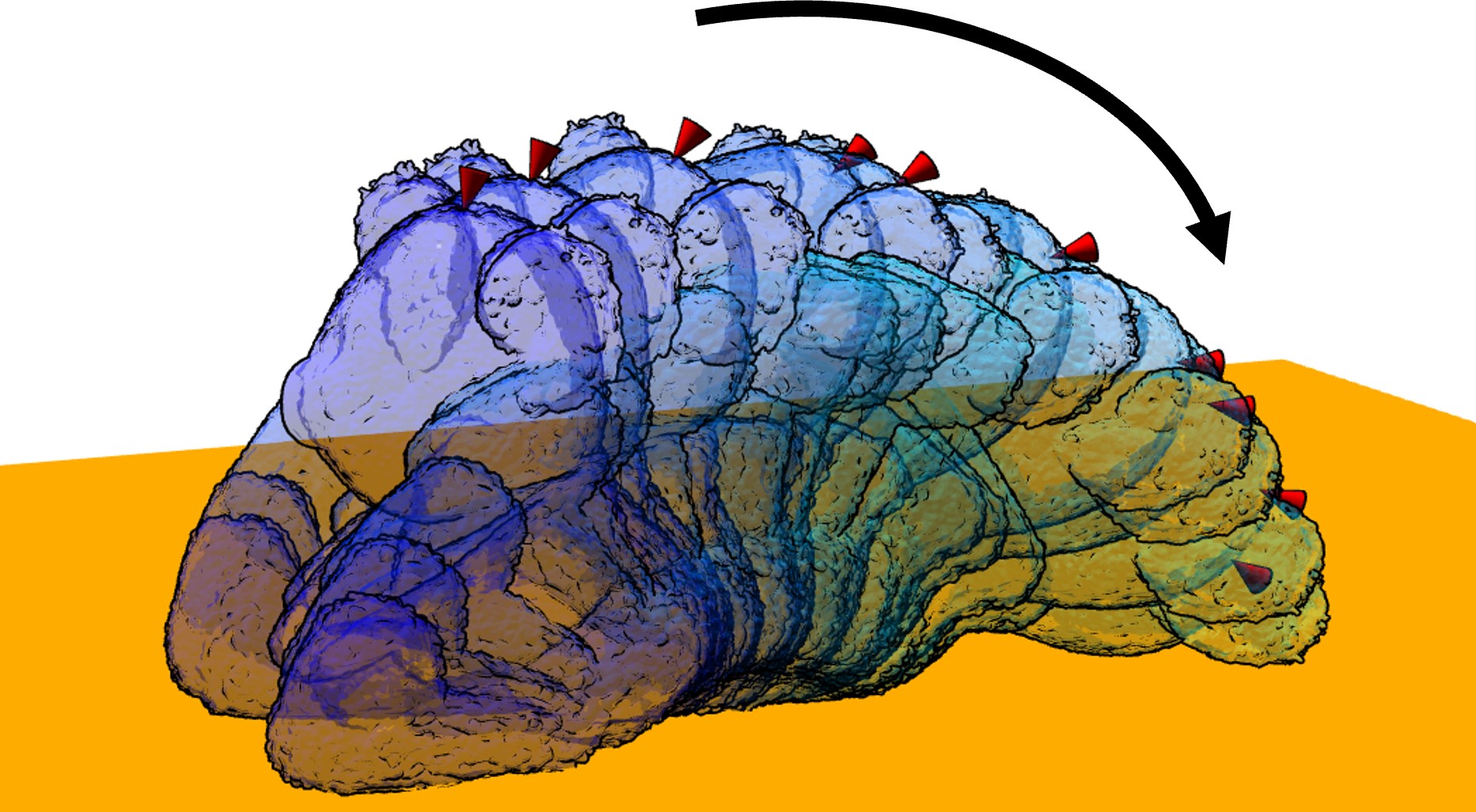} \\
         \makebox[0.3\linewidth]{\footnotesize (a) Shoe Pushing on Plane} 
         & \makebox[0.3\linewidth]{\footnotesize (b) Box Pushing on Plane} 
         & \makebox[0.3\linewidth]{\footnotesize (c) Koala Pivoting on Plane}\\
         \includegraphics[width=0.3\linewidth]{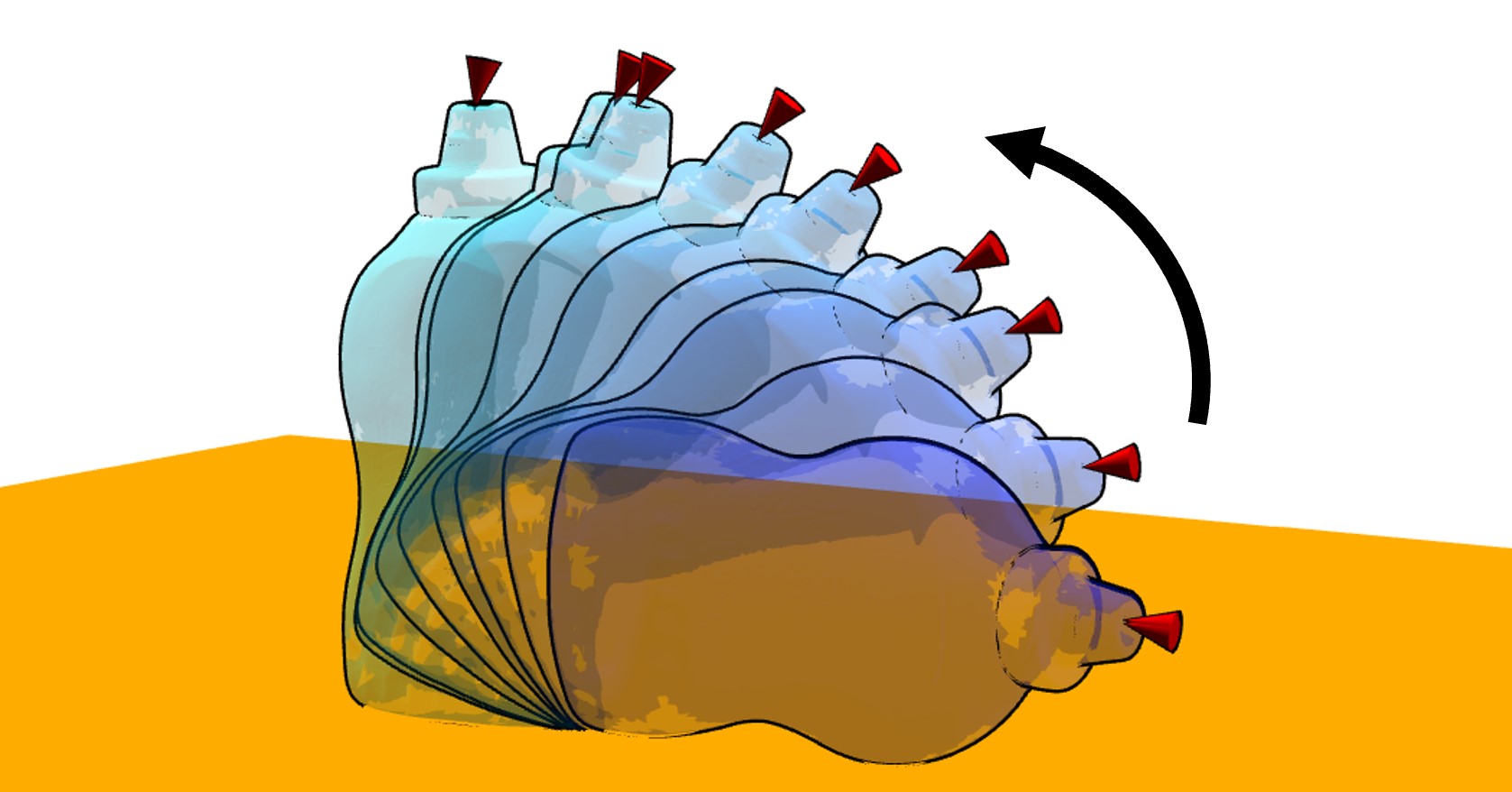}
         & \includegraphics[width=0.3\linewidth]{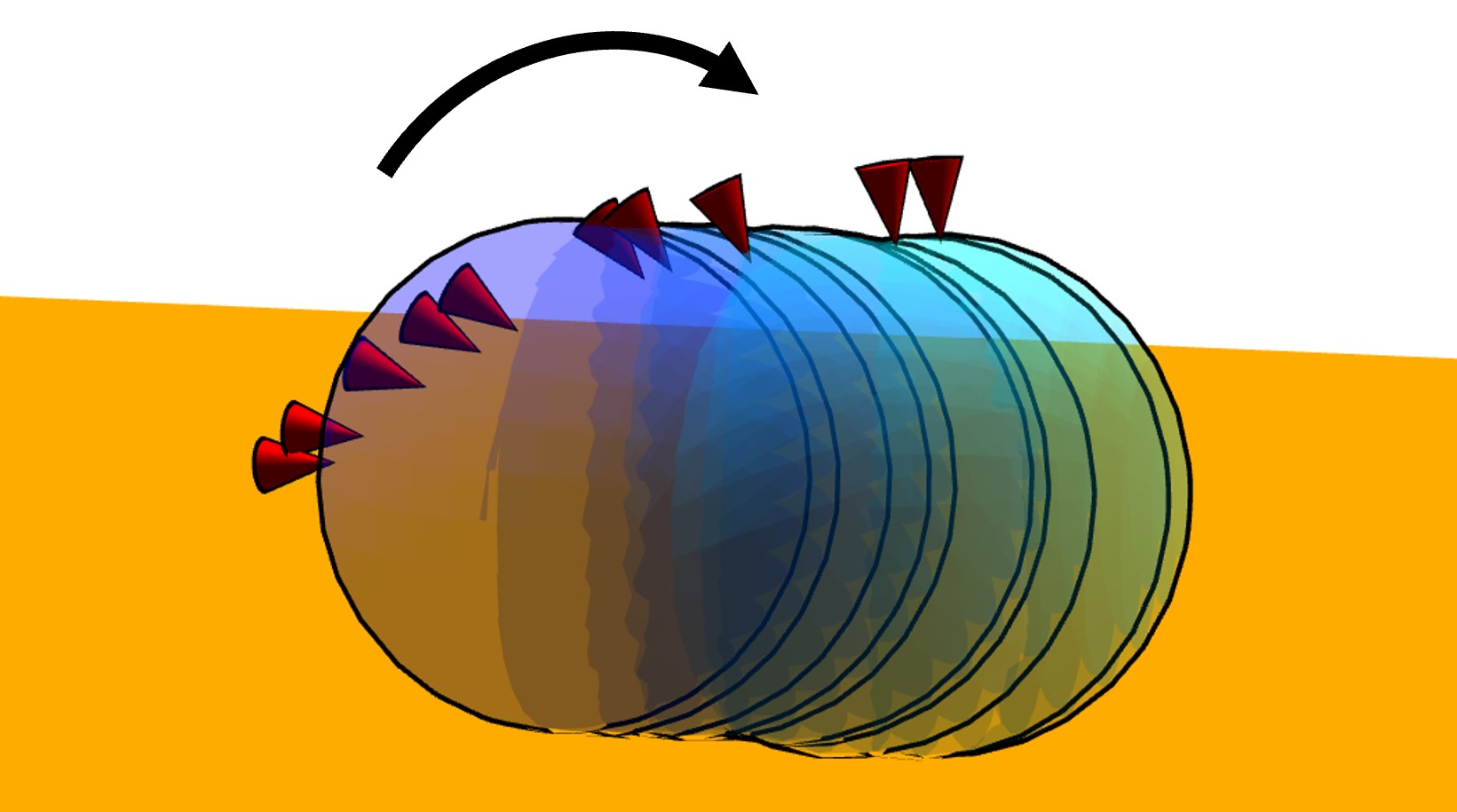}
         &  \includegraphics[width=0.3\linewidth]{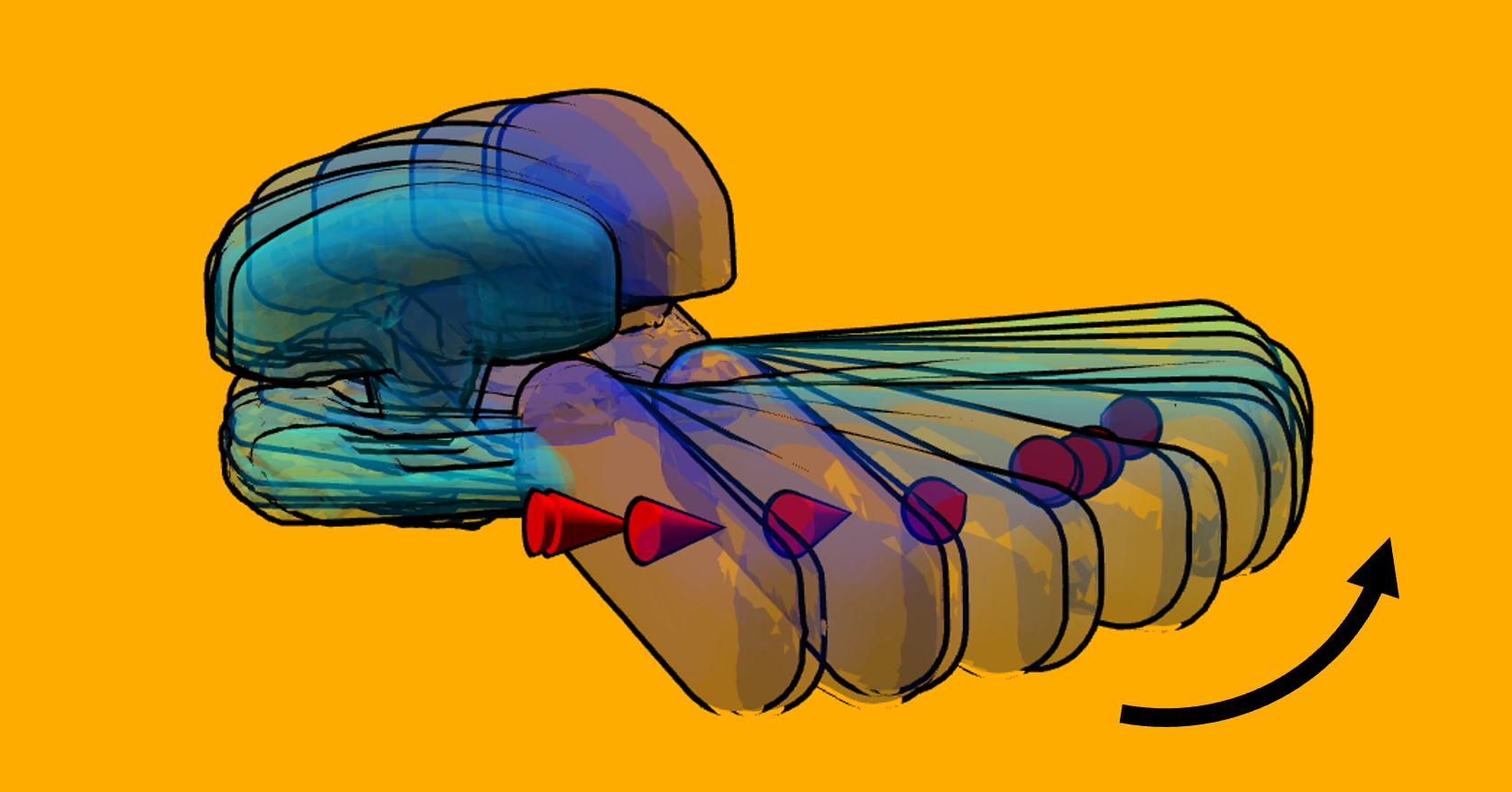} \\
         \makebox[0.3\linewidth]{\footnotesize (d) Mustard Pivoting on Plane}
         & \makebox[0.3\linewidth]{\footnotesize (e) Sphere Rolling on Plane}
         & \makebox[0.3\linewidth]{\footnotesize (f) Tool Rotating on Plane} \\
         \includegraphics[width=0.3\linewidth]{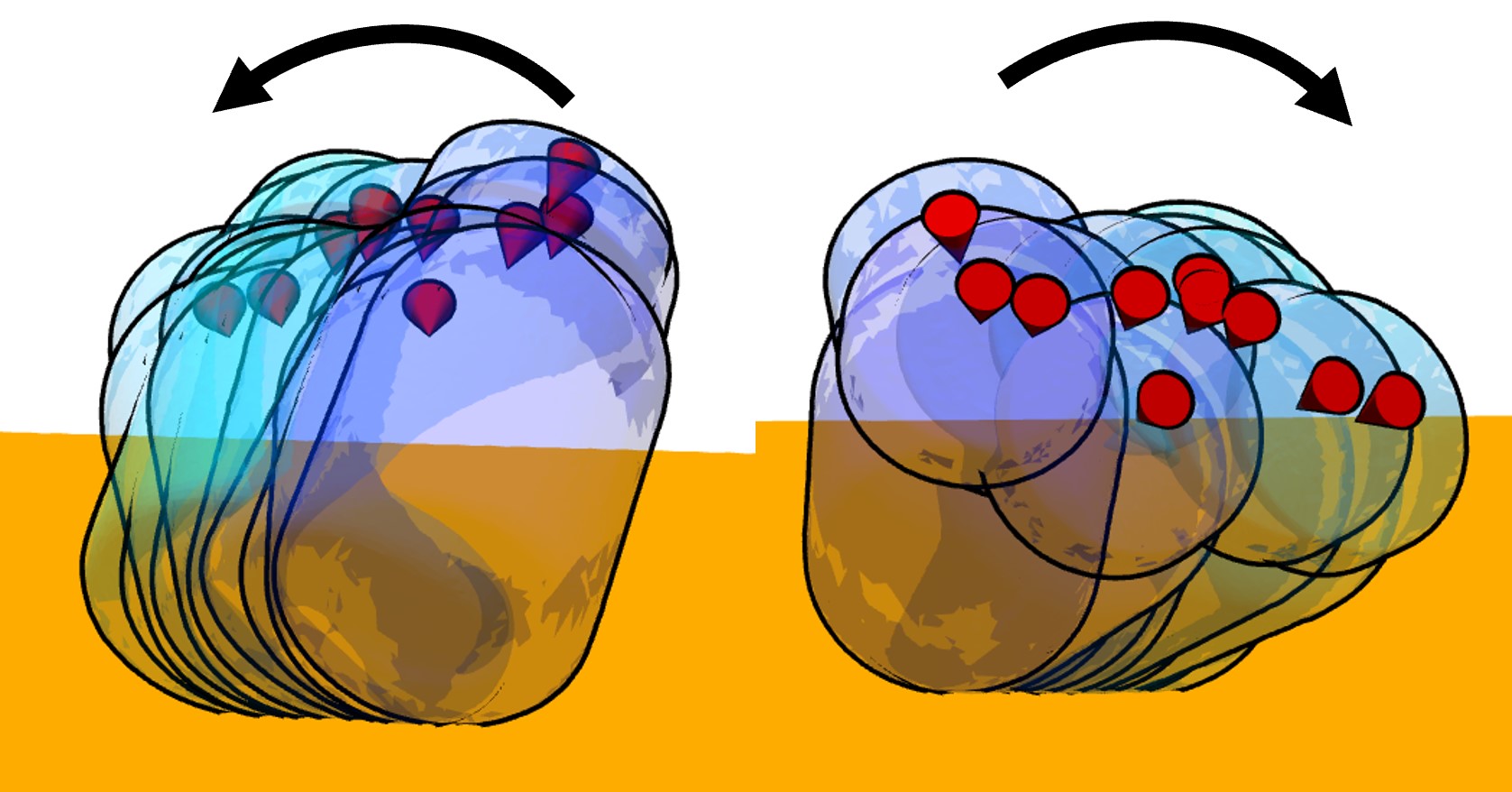}
         & \includegraphics[width=0.3\linewidth]{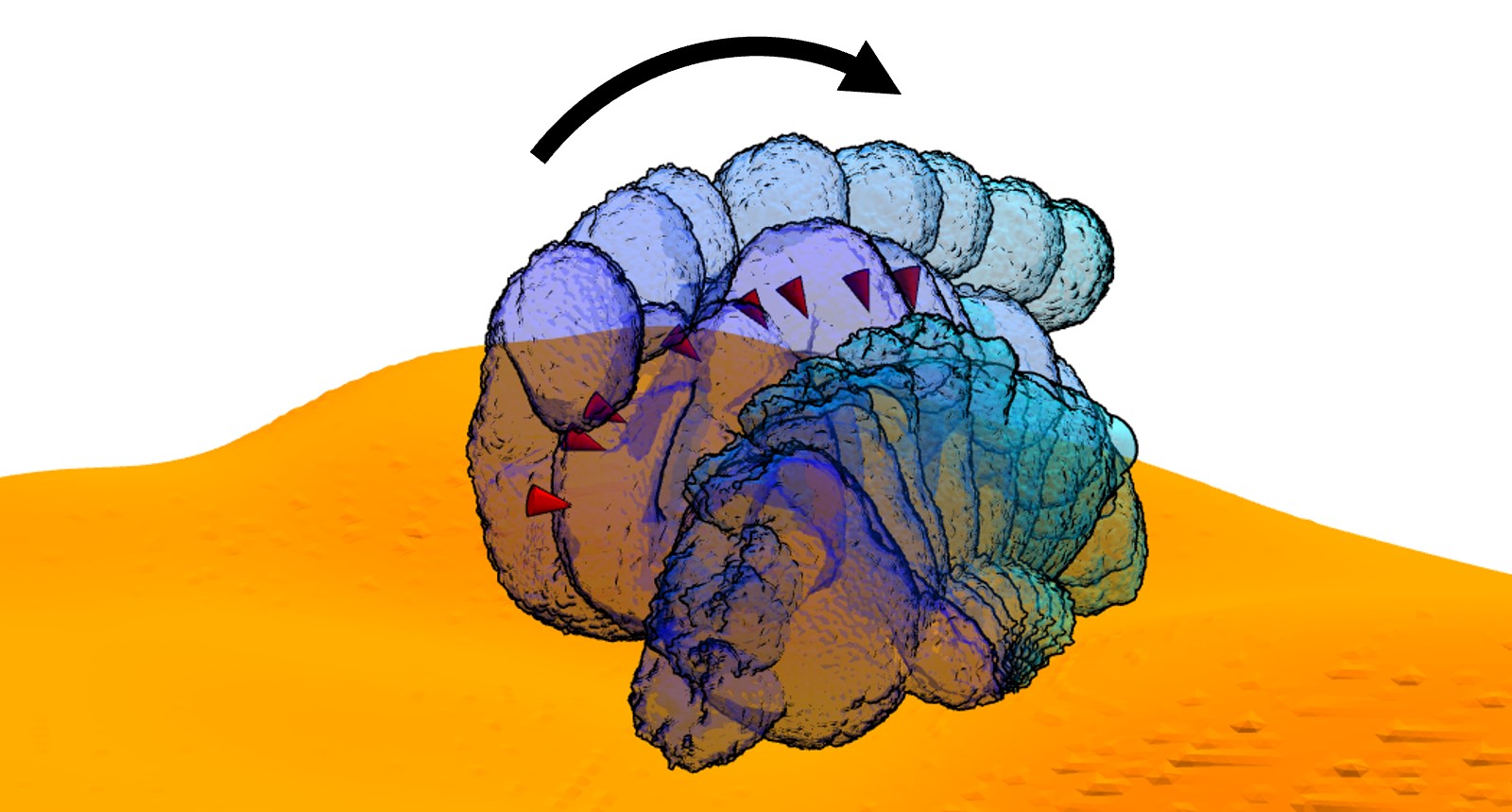}
         & \includegraphics[width=0.3\linewidth]{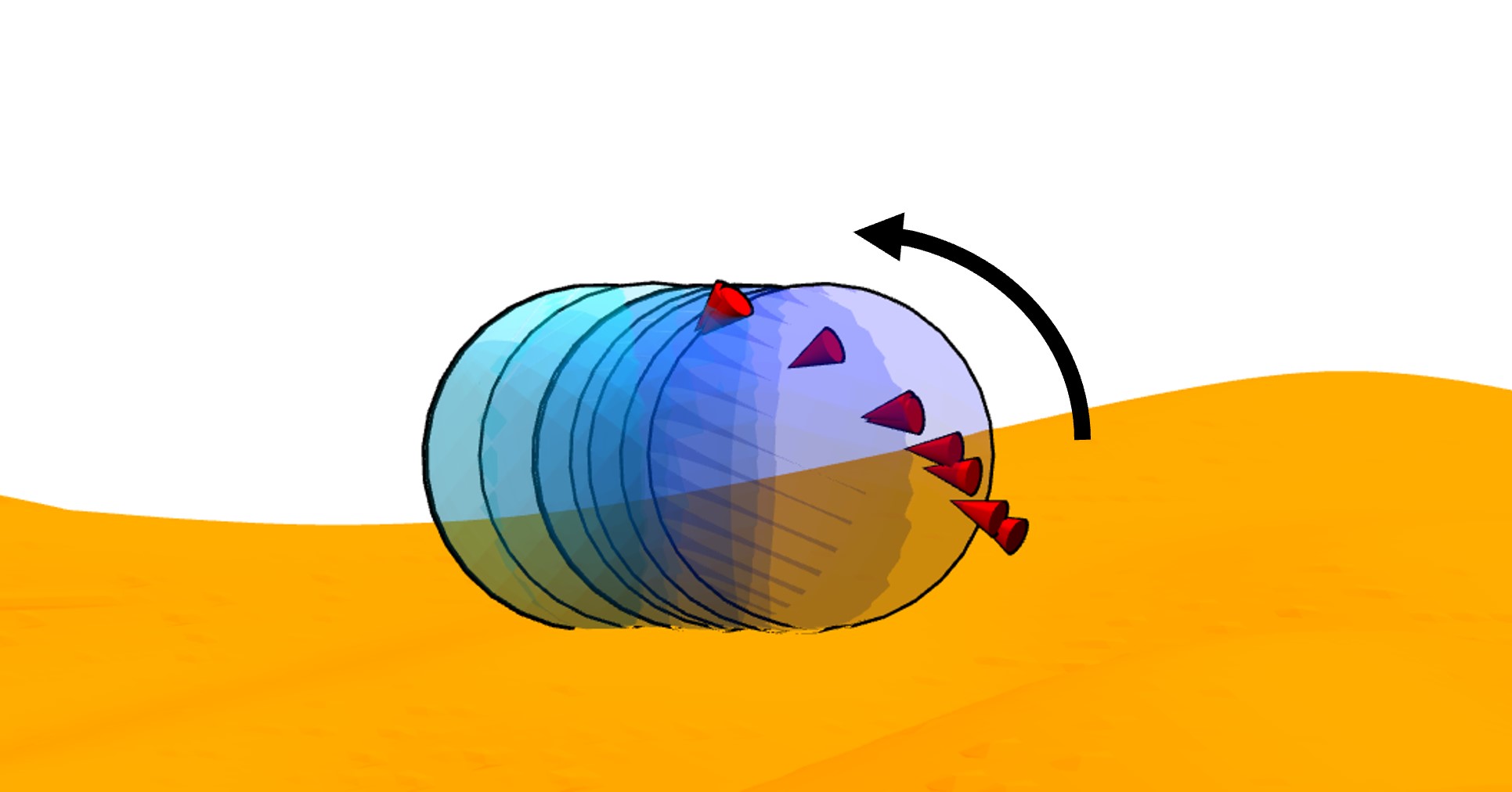}\\
         \makebox[0.3\linewidth]{\footnotesize  (g) Drug Bottle Rolling on Plane}
         & \makebox[0.3\linewidth]{\footnotesize (h) Koala Pivoting on Curved Surface}
         & \makebox[0.3\linewidth]{\footnotesize (i) Sphere Rolling on Curved Surface} \\   \includegraphics[width=0.3\linewidth]{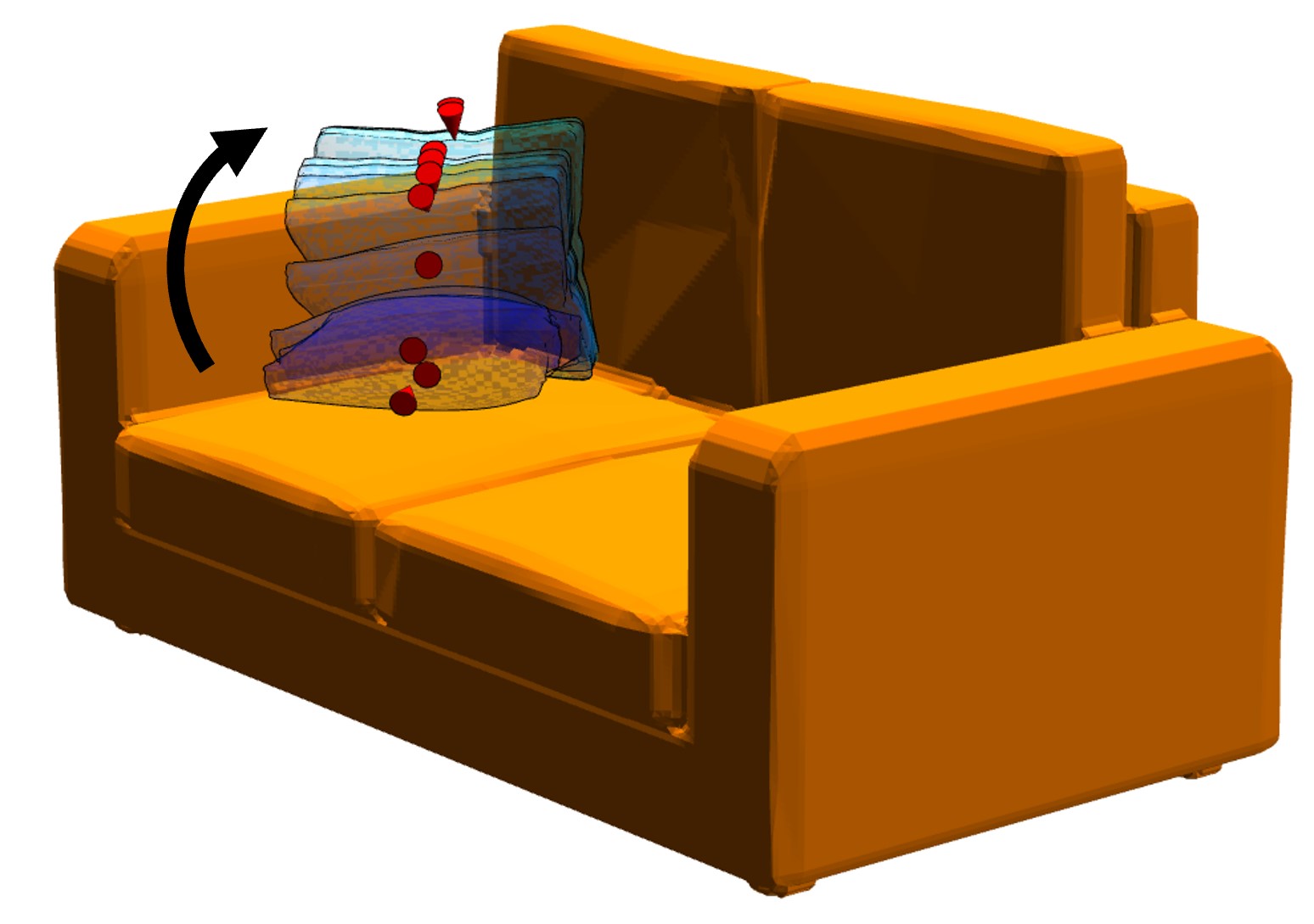}       
         & \includegraphics[width=0.3\linewidth]{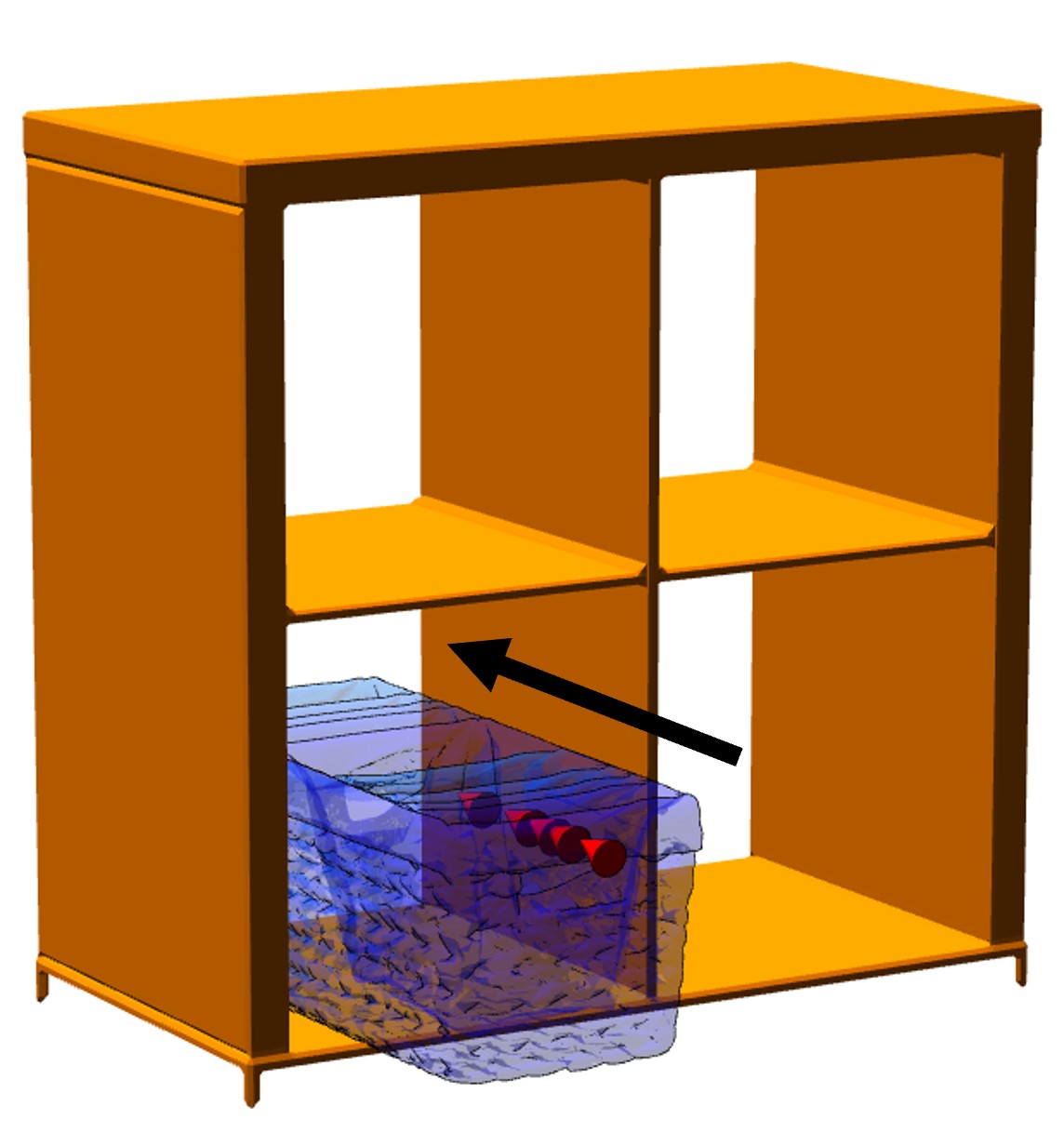}
         & \includegraphics[width=0.3\linewidth]{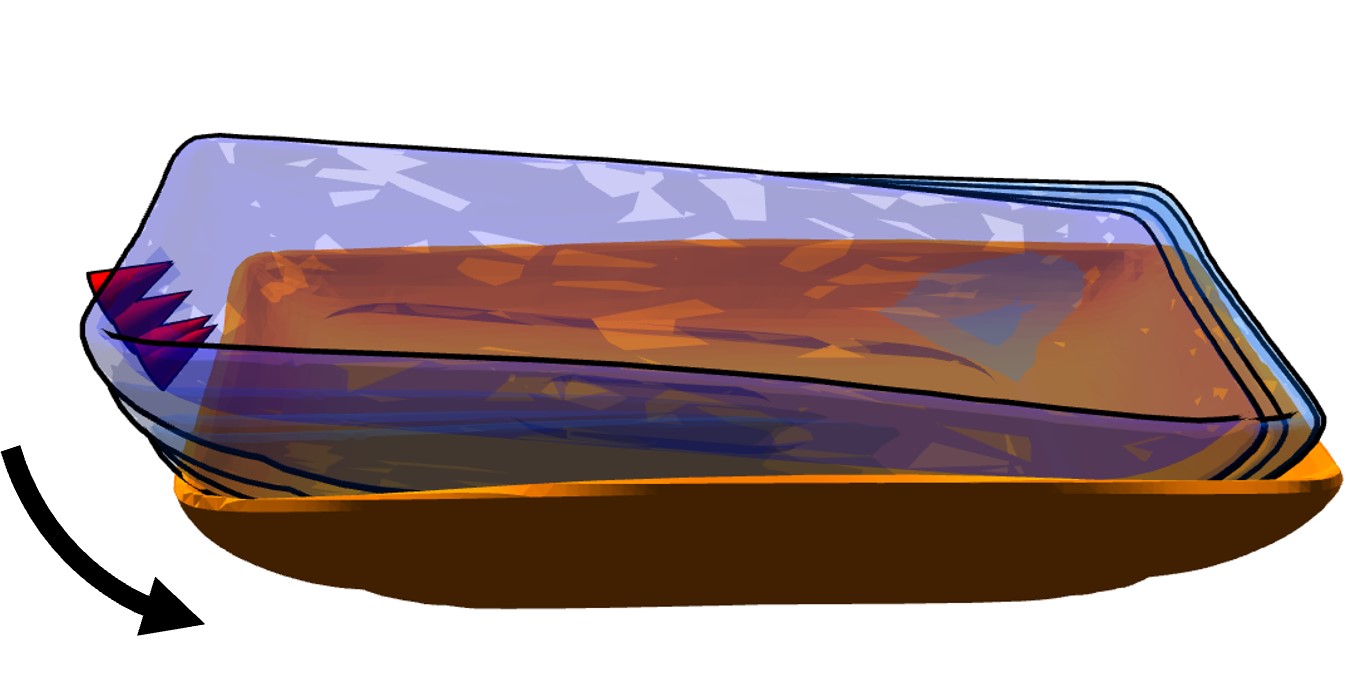} \\
         \makebox[0.3\linewidth]{\footnotesize  (j) Pillow Pivoting on Sofa}
         & \makebox[0.3\linewidth]{\footnotesize (k) Basket Pushing on Shelf}
         & \makebox[0.3\linewidth]{\footnotesize (l) Plate Sliding on Plate} \\  
         
    \end{tabular}
    \caption{Trajectories planned by STOCS for the 3D examples. Progress along the trajectory is indicated by color (dark to light). The black arrow indicates the object's movement direction. The red cone marks the contact location of the manipulator on the object.  [Best viewed in color.]}
    \label{fig:pushing_trajectories}
\end{figure*}

To demonstrate the effectiveness of STOCS in planning with high-fidelity geometric representations, we conduct experiments on ten different objects represented by dense point clouds sampled on the surface of the objects' meshes, and five different environments represented by Signed Distance Field (SDF). The SDFs are calculated offline on a grid that encloses the corresponding environment given a closed polygonal mesh of the environment, and values off of the grid vertices are approximated via trilinear interpolation.

Using STOCS, we plan for pushing, pivoting, rolling and rotating trajectories on these objects. The resulting planned trajectories are illustrated in Fig.~\ref{fig:pushing_trajectories}, while detailed information regarding the objects' geometries and the solve of the trajectories are presented in Table ~\ref{tab:1}. Like the experiment in 2D, we set $n_t=1$ and $N_s=[1e^{-2}]$ as the default parameters for TAMVO. $\Delta T=0.1\;s$, $\mu_{mnp}=1.0$ and $\mu_{env}=1.0$ are used for all the experiments in 3D. For all experiments, $T=10$ is used except in the tasks of pushing a basket on a shelf and sliding a plate on another plate, where $T=5$ is used.  To model the robot manipulatior as having a patch contact, 3 to 5 object vertices in the neighborhood of the indicated cone are allowed to be used as contact points.

\begin{table}[tbp]
\caption{Numerical results of STOCS in 3D. Number of points in the object's representation (\# Point), solve time (Time), outer iteration count (Outer iters), and average active index points for each iteration (Index points) are reported.}
\label{tab:1}
\centering
\renewcommand{\arraystretch}{1.2}
\setlength\tabcolsep{2pt}
\begin{tabular}{@{}lllllll@{}}
\toprule
Environment            & Object  & Task   & \# Point & Outer iters. & Index points & Time (s) \\
\midrule
\multirow{7}{*}{Plane} & Box     & Push   & 764      & 3            & 5.75         & 24.84    \\
                       & Shoe    & Push   & 17890    & 6            & 4.95         & 144.71    \\
                       & Koala   & Pivot  & 67359    & 4            & 5.89         & 37.93    \\
                       & Mustard & Pivot  & 8424     & 3            & 8.58         & 59.37    \\
                       & Sphere  & Roll   & 2362     & 6            & 4.39         & 94.42    \\
                       & Tool    & Rotate & 8316     & 6            & 5.09         & 99.85   \\
                       & Drug    & Roll   & 5533     & 10           & 6.35         & 319.72   \\ \hline
\multirow{2}{*}{Curve} & Koala   & Pivot  & 67359    & 9            & 13.47        & 676.01   \\
                       & Sphere  & Roll   & 2362     & 4            & 6.86         & 72.66    \\ \hline
Sofa                   & Pillow  & Pivot  & 13316    & 7            & 9.95        & 201.39    \\ \hline
Shelf                  & Basket  & Push   & 71961    & 7            & 23.10        & 421.30    \\ \hline
Plate                  & Plate   & Slide  & 67283    & 3            & 24.67        & 54.02    \\ 
\bottomrule
\end{tabular}
\end{table}

Following the initial assessments, we further evaluated the efficacy of the TAMVO alongside the SD and TS techniques through a set of comparative experiments. These experiments utilized STOCS to plan trajectories for the same set of tasks, with the primary variation being the specific oracle employed in each scenario.

Figure~\ref{fig:sf} presents the success rates of all tested Oracles. The data shows that the MVO achieves a $75\%$ success rate. In contrast, TAMVO without SD and TS exhibits worse performance than MVO; this is particularly evident in 3D scenarios where relying solely on the nearest object-to-environment point is inadequate for fulfilling the object's balance constraints. The SD and TS techniques, introduced to address this challenge, both demonstrated enhanced performance when combined with TAMVO, surpassing the success rate of TAMVO alone. Furthermore, the integration of SD and TS with TAMVO consistently achieved successful trajectory planning for all tasks.

Figure~\ref{fig:time_and_index} displays the average number of index points selected at each time step by the different Oracles assessed in our study, focusing on tasks that were successfully solved by all Oracles. As depicted in Fig.~\ref{fig:time_and_index}(a), the MVO selects a larger number of points than TAMVO and all its variations. Notably, TAMVO combined with SD and TS can successfully plan trajectories for all tasks while selecting fewer index points compared to MVO. These findings validate our hypothesis regarding TAMVO: an index point identified at time step $t$ is most valuable within a temporal vicinity of $t$. Furthermore, these results substantiate our rationale for introducing SD and TS, affirming that a localized exploration in both temporal and spatial dimensions offers a more efficient strategy than incorporating index points identified at distant time steps along the trajectory.

Figure~\ref{fig:time_and_index}(b) presents the solve times for STOCS employing various Oracles across all successful tasks. The results indicate that the quantity of index points selected by an Oracle does not necessarily correlate with the solve time. For instance, in the sphere rolling on plane task, TAMVO+SD+TS chooses a larger number of index points than TAMVO+TS, yet the solve time for TAMVO+SD+TS is much faster than that of TAMVO+TS. Similarly, in the case of the drug bottle rolling on plane task, it can be seen that both TAMVO+SD+TS and TAMVO+TS select a comparable number of index points, yet the solve time for TAMVO+TS is much faster than TAMVO+SD+TS. This phenomenon underscores that a larger number of index points does not invariably lead to longer solve time. Although TAMVO+SD+TS provides the best performance in terms of success rate, it is not guaranteed to give the best solve time for all different tasks.

\section{Conclusion and Discussion}\label{sec:conclusion}

This paper introduces Simultaneous Trajectory Optimization and Contact Selection (STOCS) to address the scaling problem in contact-implicit trajectory optimization (CITO). Through embedding CITO in an infinite programming (IP) framework to dynamically
instantiate possible contact points and contact times between
the object and environment inside the optimization loop, STOCS entends the application of CITO to a much higher number of contact points. The introduction of the Time-Active Maximum Violation Oracle along with Spatial Disturbance and Temporal Smoothing techniques for more efficient index point selection further enables STOCS to plan contact-rich manipulation trajectories with dense point clouds representation of complex-shaped objects. To the best of our knowledge, this is the first method capable of planning contact-rich manipulation trajectories with high-fidelity geometric representations in 3D.

Looking ahead, to further facilitate the practical application of the STOCS algorithm to raw object point clouds obtained from RGB-D sensors, it will be essential to devise effective pre-processing techniques for the point cloud data. This is crucial given that raw depth data typically contain a significant amount of noise. Moreover, STOCS presupposes the knowledge of specific physical properties of the object, such as mass, center of mass, and friction coefficients. However, accurately measuring these parameters in practice can be challenging, and discrepancies between the nominal values used during planning and their actual real-world counterparts may lead to failures in execution. It would be of considerable interest to explore the incorporation of robust optimization techniques into the STOCS framework to develop manipulation trajectories that maintain robustness against uncertainty in physical parameters.

\section*{Acknowledgments}
This paper is partially supported by NSF Grant \#IIS-1911087.

\bibliographystyle{IEEEtran}
\bibliography{references}

\section*{Appendix}

We discuss the detailed instantiation of the MPCC problem that needs to be solved inside STOCS for both 2D and 3D scenarios separately.

\subsection{2D}

\subsubsection{Geometry modeling}
To handle collision avoidance, we establish a semi-infinite constraint $g(q,y)$ between the object and the environment. The object is represented as a densely sampled point cloud on its surface, and the environment is represented as a signed distance field (SDF) $\psi(x):\mathbb{R}^2 \rightarrow \mathbb{R}$ which supports $O(1)$ depth lookup. Specifically, we define $g(q,y)=\psi(T_q\cdot y_p)$ with $T_q$ the object transform at configuration $q$.

\subsubsection{Friction force modeling}
The force variable of the $i^{th}$ index point at time step $t$ is $z_{t,i}=(z_{t,i}^N,z_{t,i}^+,z_{t,i}^-)$,
which is divided into the normal component $z_{t,i}^N$ and frictional components $z_{t,i}^+$ and $z_{t,i}^-$ along the tangential direction of the contact surface \cite{stewart1996implicit}. Given the normal vector $n_{t,i}^{N_{z}}$ along the outward surface normal of the environment, and $n_{t,i}^{+_{z}}$ and $n_{t,i}^{-_{z}}$, the overall force applied at index point $y_{t,i}$ is $z_{t,i} = z_{t,i}^{N} n_{t,i}^{N_{z}} +  z_{t,i}^+ n_{t,i}^{+_{z}} +  z_{t,i}^- n_{t,i}^{-_{z}}$.  Each of the components of $z_{t,i}$ is required to be non-negative, and given the friction coefficient $\mu_{env}$, the friction cone constraint is given by $\mu_{env} z_{t,i}^N - (z_{t,i}^++z_{t,i}^-) \geq 0$.

Similarly, for the $i^{th}$ contact between the robot and the object, we define the force variable at time step $t$ to be $u_{t,i}=(u_{t,i}^N,u_{t,i}^+,u_{t,i}^-)$. Besides, the normal vector $n_{t,i}^{N_{u}}$ along the inward surface normal of the object, and the tangential vectors $n_{t,i}^{+_{u}}$ and $n_{t,i}^{-_{u}}$ which are perpendicular to the normal vector are defined in the object's local frame. 

We adopt a heuristic to reduce the number of constraints in the complementarity condition and accelerate solve times. By only applying complementarity to the normal component of force variables,  $g(q_t,y_{t,i})\cdot z_{t,i}^N = 0$, we reduce the number of distance complementarity constraints from $\sum_0^{T} 3|\tilde{Y}_t|$ to $\sum_0^{T} |\tilde{Y}_t|$, and the problem is unchanged because $z_{t,i}^N=0 \Rightarrow z_{t,i}^+=0,\;z_{t,i}^-=0$ due to the friction cone constraint. 

\subsubsection{Force and torque balance}
We establish an equality constraint on the object's force and torque balance. It requires the joint force and torque exerted by the robot, the environment, and the gravity on the object to be $0$ if we assume quasistatic, and to be $M\dot{v}$ if we assume quasidynamic, where $M$ is the inertia matrix of the object $\mathcal{O}$ and $v$ is the velocity of the object.

\subsubsection{Instantiation of MPCC problem}

With these definitions, the nonlinear programming problem $P_k$ that needs to be solved at the $k^{th}$ iteration of STOCS is the following:

\begin{small}
\begin{subequations}
\begin{align}
\label{pb:5}
\min_{q,v,u,z} &\; \tilde{f}(q,v,u,z)\\
s.t. &\;q_0\in\mathcal{Q}_{init},q_T \in\mathcal{Q}_{goal} \\
&\; q_t - q_{t+1} + v_{t+1}\Delta t = 0 \quad \forall t \in \timeSet \setminus \{T\}\\
&\; q_t\in\mathcal{Q},\; u_t\in\mathcal{U},v_t\in \mathcal{V} \quad \forall t \in \timeSet\\ 
&\; \mu_{mnp} u_{t,i}^N - (u_{t,i}^+ + u_{t,i}^-) \geq 0 \quad \forall i \in \mathcal{C}, \quad \forall t \in \timeSet \\
&\; 0 \leq z_{t,i}^N \perp g(q_t,y_{t,i}) \geq 0 \quad \forall i \in \mathcal{I}_t, \quad \forall t \in \timeSet\\
&\; z_{t,i}^+,z_{t,i}^- \geq 0 \quad \forall i \in \mathcal{I}_t, \quad \forall t \in \timeSet\\
&\; \mu_{env} z_{t,i}^N - (z_{t,i}^++z_{t,i}^-) \geq 0 \quad \forall i \in \mathcal{I}_t, \quad \forall t \in \timeSet\\
&\;\gamma_{t,i}+\psi_i(q_t,v_t) \geq 0, \quad \gamma_{t,i}-\psi_i(q_t,v_t) \geq 0 \nonumber \\ 
&\; \forall i \in \mathcal{I}_t, \quad \forall t \in \timeSet\\
&\;(\mu z_{t,i}^N - (z_{t,i}^++z_{t,i}^-))^T \gamma_{t,i} = 0 \quad \forall i \in \mathcal{I}_t, \quad \forall t \in \timeSet\\
&\;(\gamma_{t,i}+\psi_i(q_t,v_t))^T  z_{t,i}^+ = 0 \quad \forall i \in \mathcal{I}_t, \quad \forall t \in \timeSet\\
&\;(\gamma_{t,i}-\psi_i(q_t,v_t))^T  z_{t,i}^- = 0 \quad \forall i \in \mathcal{I}_t, \quad \forall t \in \timeSet\\
&\; z_{t,i} = z_{t,i}^{N}\cdot n_{t,i}^{N_z} + z_{t,i}^+\cdot n_{t,i}^{+_{z}} + z_{t,i}^-\cdot n_{t,i}^{-_{z}} \nonumber \\
&\; \forall i \in \mathcal{I}_t, \quad \forall t \in \timeSet \\
&\; u_{t,i} = u_{t,i}^{N} \cdot n_{t,i}^{N_{u}}+u_{t,i}^{+} \cdot n_{t,i}^{+_{u}}+u_{t,i}^{-} \cdot n_{t,i}^{-_{u}} \nonumber \\
&\; \forall i \in \mathcal{C}, \quad \forall t \in \timeSet \\
&\;mg+ \sum_i^{\mathcal{C}} u_{t,i} + \sum_i^{\mathcal{I}_t} z_{t,i} = 0 \; \text{or} \; M_{trans}\dot{v}_{trans} \quad \forall t \in \timeSet \\
&\; \sum_i^{\mathcal{C}} (q_t \cdot (c^{mnp,i}-CoM)) \times u_{t,i} + \nonumber \\ 
&\;\sum_i^{\mathcal{I}_t} (q_t\cdot(y_{t,i} - CoM)) \times z_{t,i} = 0 \; \text{or} \; M_{rot}\dot{v}_{rot} \quad \forall t \in \timeSet
\end{align}
\vspace{-10pt}
\end{subequations}
\end{small}

where $\timeSet=\{0,\cdots,T\}$,   $\mathcal{I}_t = \{0,\ldots,|\tilde{Y}_t|\}$, $\psi_i(q_t,v_t)$ is the relative tangential velocity at contact point $y_{t,i}$. $\mathcal{C}=\{0,\cdots,|c^{mnp}|\}$, $c^{mnp,i}$ is the $i^{th}$ contact point between the robot and the object in the object's frame, $CoM$ is the center of mass of the object, $v_{trans}$ is the translational velocity of the object, $M_{trans}$ is the translational part of the inertia matrix of the object with appropriate dimension, and $v_{rot}$ and $M_{rot}$ are the corresponding terms for the rotational part.

\subsection{3D}

\subsubsection{Geometry modeling}
To handle collision avoidance, like in the 2D scenario, we establish a semi-infinite constraint $g(q,y)$ between the object and the environment. The object is represented as a densely sampled point cloud on its surface, and the environment is represented as a signed distance field (SDF) $\psi(x):\mathbb{R}^3 \rightarrow \mathbb{R}$ which supports $O(1)$ depth lookup. We define $g(q,y)=\psi(T_q\cdot y_p)$ with $T_q$ the object transform at configuration $q$.

\subsubsection{Friction force modeling}
In 3D scenario, following \cite{posa2014direct}, to preserve the MPCC structure of the resulting optimization problem, we use a polyhedral approximation of the friction cone \cite{stewart1996implicit}. The force variable of the $i^{th}$ index point at time step $t$ is $z_{t,i}=(z^N_{t,i},z^{D1}_{t,i},\cdots,z^{Dd}_{t,i})$, which is expressed in a reference frame with $z^N_{t,i}$ normal to the contact surface, and $z^{D1}_{t,i}, \cdots, z^{Dd}_{t,i}$ tangent to the contact surface. The convex hull of the unit vectors along the directions of $z^{D1}_{t,i}, \cdots, z^{Dd}_{t,i}$ in $\mathbb{R}^2$ forms the polyhedral approximation. The normal vector $n_{t,i}^{N_{z}}$ is along the outward surface normal of the environment, and $n_{t,i}^{D1_{z}}, \cdots,  n_{t,i}^{Dd_{z}}$ are perpendicular to $n_{t,i}^{N_{z}}$. Each of the components of $z_{t,i}$ is required to be non-negative, and given the friction coefficient $\mu_{env}$, the friction cone constraint is given by $\mu_{env} z_{t,i}^N - (z_{t,i}^{D1}+\cdots+z_{t,i}^{Dd}) \geq 0$.

Similarly, for the $i^{th}$ contact between the robot and the object, we define the force variable at time step $t$ to be $u_{t,i}=(u_{t,i}^N,u^{D1}_{t,i},\cdots,u^{Dd}_{t,i})$. The normal vector $n_{t,i}^{N_{u}}$ is along the inward surface normal of the object, and $n_{t,i}^{D1_{u}}, \cdots,  n_{t,i}^{Dd_{u}}$ are perpendicular to $n_{t,i}^{N_{u}}$. 

The same heuristic is used to reduce the number of constraints in the complementarity condition and accelerate solve times. By only applying complementarity to the normal component of force variables,  $g(q_t,y_{t,i})\cdot z_{t,i}^N = 0$, we reduce the number of distance complementarity constraints from $\sum_0^{T} (d+1)|\tilde{Y}_t|$ to $\sum_0^{T} |\tilde{Y}_t|$, and the problem is unchanged because $z_{t,i}^N=0 \Rightarrow z_{t,i}^{Dj}=0 \; \forall j \in \{1,\cdots,d\}$ due to the friction cone constraint. 

\subsubsection{Instantiation of MPCC problem}

With these definitions the nonlinear programming problem $P_k$ that needs to be solved at the $k^{th}$ iteration for this pivoting task is the following:

\begin{small}
\begin{subequations}
\begin{align}
\label{pb:5}
\min_{q,v,u,z} &\; \tilde{f}(q,v,u,z)\\
s.t. &\;q_0\in\mathcal{Q}_{init},q_T \in\mathcal{Q}_{goal} \\
&\; q_t - q_{t+1} + v_{t+1}\Delta t = 0 \quad \forall t \in \timeSet \setminus \{T\}\\
&\; q_t\in\mathcal{Q},\; u_t\in\mathcal{U},v_t\in \mathcal{V} \quad \forall t \in \timeSet\\ 
&\; \mu_{mnp} u_{t,i}^N - \sum_{j}u_{t,i}^{D_j} \geq 0 \nonumber \\
&\;  \forall i \in \mathcal{C}, \quad \forall j \in \{1,\cdots,d\}, \quad \forall t \in \timeSet \\
&\; 0 \leq z_{t,i}^N \perp g(q_t,y_{t,i}) \geq 0 \quad \forall i \in \mathcal{I}_t, \quad \forall t \in \timeSet\\
&\; z_{t,i}^{D_j} \geq 0 \quad \forall i \in \mathcal{I}_t, \quad \forall j \in \{1,\cdots,d\}, \quad \forall t \in \timeSet\\
&\; \mu_{env} z_{t,i}^N - \sum_{j}z_{t,i}^{D_j} \geq 0 \nonumber \\
&\; \quad \forall i \in \mathcal{I}_t, \quad \forall j \in \{1,\cdots,d\}, \quad \forall t \in \timeSet\\
&\;\gamma_{t,i}+\psi(q_t,v_t,y_{t,i}) \cdot n_{t,i}^{D_j} \geq 0 \nonumber \\
&\;\quad \forall i \in \mathcal{I}_t, \quad \forall j \in \{1,\cdots,d\}, \quad \forall t \in \timeSet\\
&\;(\mu z_{t,i}^N - \sum_{j} z_{t,i}^{D_j}) \gamma_{t,i} = 0 \nonumber \\
&\;\forall i \in \mathcal{I}_t, \quad \forall j \in \{1,\cdots,d\}, \quad \forall t \in \timeSet\\
&\;(\gamma_{t,i}+\psi(q_t,v_t,y_{t,i})\cdot n_{t,i}^{D_j})  z_{t,i}^{D_j} = 0 \nonumber \\
&\;\forall i \in \mathcal{I}_t, \quad \forall j \in \{1,\cdots,d\}, \quad \forall t \in \timeSet\\
&\; z_{t,i} = z_{t,i}^N n_{t,i}^{N_{z}} +  \sum_{j}z_{t,i}^{Dj} n_{t,i}^{Dj_{z}} \nonumber \\
&\; \quad \forall i \in \mathcal{I}_t, \quad \forall j \in \{1,\cdots,d\}, \quad \forall t \in \timeSet \\
&\; u_{t,i} = u_{t,i}^{N} n_{t,i}^{N_{u}} + \sum_{j}u_{t,i}^{Dj} n_{t,i}^{Dj_{u}} \nonumber \\
&\; \quad \forall i \in \mathcal{C}, \; \forall j \in \{1,\cdots,d\}\quad \forall t \in \timeSet \\
&\;mg+ \sum_i^{C}u_{t,i} + \sum_i^{\mathcal{I}_t} z_{t,i} = 0 \; \text{or} \; M_{trans}\dot{v}_{trans} \quad \forall t \in \timeSet \\
&\; \sum_i^{\mathcal{C}}(q_t \cdot (c^{mnp,i}-CoM)) \times u_{t,i} + \nonumber \\ 
&\;\sum_i^{\mathcal{I}_t} (q_t\cdot(y_{t,i} - CoM)) \times z_{t,i} = 0 \; \text{or} \; M_{rot}\dot{v}_{rot}\quad \forall t \in \timeSet
\end{align}
\vspace{-10pt}
\end{subequations}
\end{small}

where $\timeSet=\{0,\cdots,T\}$, $\mathcal{I}_t = \{0,\ldots,|\tilde{Y}_t|\}$, $\psi(q_t,v_t,y_{t,i})$ is the relative tangential velocity at contact point $y_{t,i}$. $\mathcal{C}=\{0,\cdots,|c^{mnp}|\}$, $c^{mnp,i}$ is the $i^{th}$ contact point between the robot and the object in the object's frame, $CoM$ is the center of mass of the object, $v_{trans}$ is the translational velocity of the object, $M_{trans}$ is the translational part of the inertia matrix of the object with appropriate dimension, and $v_{rot}$ and $M_{rot}$ are the corresponding terms for the rotational part.

\end{document}